\documentclass[10pt,twocolumn,letterpaper]{article}

\usepackage[pagenumbers]{paper}

\usepackage{capt-of}

\definecolor{linkblue}{rgb}{0.21,0.49,0.74}
\usepackage[pagebackref,breaklinks,colorlinks,allcolors=linkblue]{hyperref}
\usepackage{url}
\usepackage{tikz}
\usepackage{xcolor}
\usepackage{multirow}
\usetikzlibrary{arrows.meta}
\usetikzlibrary{shapes.geometric, arrows.meta, patterns, calc}
\usetikzlibrary{fit, backgrounds}

\usepackage{amsmath,amssymb,booktabs}
\usepackage{algorithm}
\usepackage{algpseudocode}
\usepackage{mathtools}
\usepackage[normalem]{ulem}

\newtheorem{prop}{Proposition}

\definecolor{mpcblue50}{RGB}{230,241,251}
\definecolor{mpcblue600}{RGB}{24,95,165}
\definecolor{mpcblue800}{RGB}{12,68,124}
\definecolor{mpcteal50}{RGB}{225,245,238}
\definecolor{mpcteal600}{RGB}{15,110,86}
\definecolor{mpcteal800}{RGB}{8,80,65}
\definecolor{mpcgray50}{RGB}{241,239,232}
\definecolor{mpcgray600}{RGB}{95,94,90}
\definecolor{mpcgray800}{RGB}{68,68,65}
\definecolor{mpcred50}{RGB}{252,235,235}
\definecolor{mpcred600}{RGB}{163,45,45}
\definecolor{mpcred800}{RGB}{121,31,31}
\definecolor{mpcamber50}{HTML}{FFFBEB}
\definecolor{mpcamber600}{HTML}{D97706}
\definecolor{mpcamber800}{HTML}{92400E}

\title{Skill-JEPA: Learning Planner-Aligned Action Spaces for World Models}

\author{Tom\'as Couso Coddou\\
Centro Nacional De Inteligencia Artificial\\
{\tt\small tomas.couso@cenia.cl}
\and
Christ Devia\\
Centro Nacional De Inteligencia Artificial\\
{\tt\small christ.devia@cenia.cl}
\and
Rex G. Liu\\
Centro Nacional De Inteligencia Artificial\\
{\tt\small rex.liu@cenia.cl}
\and
Rodrigo Toro Icarte\\
Pontificia Universidad Católica de Chile\\
{\tt\small rntoro@uc.cl}
}

\begin{document}
\twocolumn[{%
\renewcommand\twocolumn[1][]{#1}%
\maketitle
}]

\begin{abstract}
World models based on Joint-Embedding Predictive Architectures (JEPAs) learn from pixels to predict the outcome of actions in a latent space. To reach a goal, a planner samples many candidate action sequences, uses the model to predict where each one leads, and executes the best one.
However, standard planners draw their candidates from a Gaussian,
while the model has only seen the structured, state-dependent actions of its training data. Its predictions for such candidates are unreliable, and planning fails more often as goals move farther away. We propose Skill-JEPA, which learns an action space aligned with the planner. A variational autoencoder conditioned on the current state encodes short action sequences into latent skills with a standard Gaussian prior, and decodes them back into actions. The world model predicts the effect of skills rather than raw actions, and the planner searches over skills starting from that same Gaussian, so its candidates resemble the model's training data. Skill-JEPA adds a single module, trained end-to-end with the world model, and leaves the planner untouched, with no subgoals, hierarchy, or hand-designed skills. Across three JEPA world models and four benchmarks, average success on goals 80 steps away rises from 42\% with raw actions to 62\%; the gains grow with goal distance and hold across four planners. On AntMaze, a quadruped locomotion task, success on goals two maze cells away rises from 0.4\% to 38\%. This suggests that part of the difficulty of reaching distant goals lies in the action space, and can be removed without changing the planner.%
\end{abstract}

\section{Introduction}
\label{sec:intro}

A world model predicts how an environment responds to an agent's actions~\cite{ha2018worldmodels}. With such a model, the agent can plan before it acts~\cite{Sutton1998}: to reach a goal, it imagines the outcome of many candidate action sequences and executes the most promising one. World models based on Joint-Embedding Predictive Architectures (JEPAs)~\cite{lecun-autonomous-machine-intelligence-2022} make these predictions in a learned latent space rather than in pixel space, and are trained on offline trajectories without rewards~\cite{zhou2025dinowmworldmodelspretrained, assran2025vjepa2selfsupervisedvideo, sobal2025learningrewardfreeofflinedata, maes_lelidec2026lewm, ivashkov2026sensorimotorworldmodelsperception}. Nothing in this recipe is tied to a task. Once trained, the same model can pursue any goal given as an image, using an off-the-shelf sampling-based planner such as the Cross-Entropy Method (CEM)~\cite{Rubinstein1999}. Planning in this way has been found to generalize better than goal-conditioned policies learned from the same data~\cite{sobal2025learningrewardfreeofflinedata}.

\begin{figure*}
    \centering
    \raisebox{-0.5\height}{\includegraphics[height=0.21\textwidth]{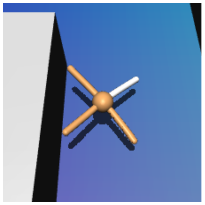}}\hspace{2em}%
    \raisebox{-0.5\height}{\includegraphics[height=0.27\textwidth]{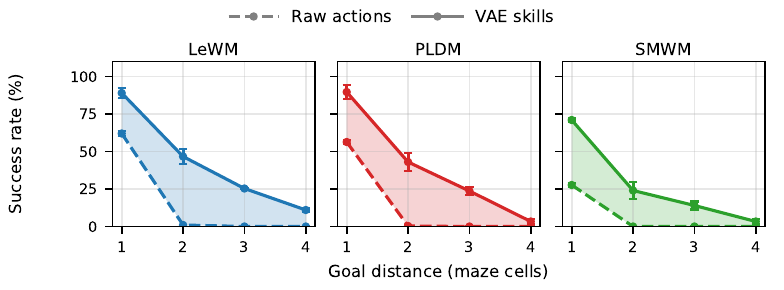}}
    \caption{\textbf{The same planner reaches farther goals when it searches over learned skills instead of raw actions.}
    In OGBench-AntMaze~\citep{park2025ogbenchbenchmarkingofflinegoalconditioned} (left), a quadruped with eight actuated joints must reach goals up to four maze cells away.
    For three JEPA world models (right), we plan over raw actions (dashed) or over the skills learned by Skill-JEPA (solid).
    The planner, the data, and the encoder and predictor architectures are identical; only the action space differs between Raw actions and VAE skills.
    Averaged over the three models, success at two cells is 0.4\% with Raw actions and 38\% with Skills, which still succeed occasionally at four cells (6\%).}
    \label{fig:teaser}
\end{figure*}

The model and the planner, however, do not work with the same actions. The model learns from the actions recorded in its training data, which are structured and depend on the state: a walking robot moves its legs in coordination, and the right movement depends on its current pose. The planner knows none of this. CEM, like the other planners commonly used with these models, draws candidates from a Gaussian, scores them with the model, and refines its search around the best ones.
For the quadruped in Figure~\ref{fig:teaser}, a Gaussian candidate sends unrelated random commands to its eight joints at every step: a twitch, not a gait.

The model is therefore asked to predict the outcome of action sequences unlike those it has seen, and its predictions for them are unreliable. Because the planner chooses candidates by their predicted outcome, these errors mislead it, and they accumulate as the model is rolled out toward more distant goals. In our experiments, planning over raw actions reaches on average 87\% of the goals that are 20 steps away, but only 42\% of those 80 steps away. This gap between the actions a model is trained on and those it is queried with has been documented for latent world models~\cite{agrawal2025closing, you2026control}. It echoes a long-standing observation in model-based reinforcement learning: optimizers exploit the errors of learned models~\cite{yu2020mopo, janner2022diffuser}. %

Existing remedies move one side toward the other. Some bring the model to the planner, by fine-tuning it on trajectories produced by planning or on adversarially perturbed data~\cite{agrawal2025closing}. Others bring the planner to the data, by training a policy that proposes candidates~\cite{argenson2021mbop, chahe2026pijepa, hegde2026lewam}. Yet others shorten the problem itself, with a learned model that proposes subgoals~\cite{masip2026ffjepa} or a hierarchy of world models that plan at several time scales~\cite{zhang2026hierarchicalplanninglatentworld, zhang2026hjepaendtoendlearninghierarchical}. These approaches help, but each adds a learned component, a training stage or a level of planning, and gives up part of the simplicity that makes the recipe attractive.

\begin{figure*}[ht]
\centering
\definecolor{panelA}{RGB}{243,247,252}\definecolor{panelAborder}{RGB}{190,208,228}
\definecolor{panelB}{RGB}{252,248,240}\definecolor{panelBborder}{RGB}{226,208,176}
\begin{minipage}[t]{0.495\textwidth}
\centering
\resizebox{\linewidth}{!}{%
\begin{tikzpicture}[
  >=Stealth,
  x=0.7cm, y=0.8cm,
  font=\small,
  obs/.style={
    rectangle, rounded corners=5pt,
    fill=mpcgray50, draw=mpcgray600, line width=0.4pt,
    minimum width=1.6cm, minimum height=0.7cm, inner sep=4pt
  },
  enc/.style={
    rectangle, rounded corners=5pt,
    fill=mpcblue50, draw=mpcblue600, line width=0.4pt,
    minimum width=1.6cm, minimum height=0.7cm, inner sep=4pt
  },
  lat/.style={
    rectangle, rounded corners=5pt,
    fill=mpcteal50, draw=mpcteal600, line width=0.4pt,
    minimum width=1.6cm, minimum height=0.7cm, inner sep=4pt
  },
  pred/.style={
    rectangle, rounded corners=5pt,
    fill=mpcblue50, draw=mpcblue600, line width=0.4pt,
    minimum width=2.2cm, minimum height=0.7cm, inner sep=4pt
  },
  zhat/.style={
    rectangle, rounded corners=5pt,
    fill=mpcblue50, draw=mpcblue600, line width=0.4pt,
    dashed,
    minimum width=1.7cm, minimum height=0.7cm, inner sep=4pt
  },
  mse/.style={
    rectangle, rounded corners=4pt,
    fill=mpcred50, draw=mpcred600, line width=0.4pt,
    minimum width=1.1cm, minimum height=0.55cm, inner sep=3pt
  },
  junction/.style={circle, fill=mpcteal600, inner sep=0pt, minimum size=2.4pt},
  bridge/.style={preaction={draw=panelA, line width=2.4pt}}
]

\node[obs] (St)    at (-4.4, -6.4) {\textcolor{mpcgray800}{$s_t$}};
\node[enc] (Eleft) at (-4.4, -4.6) {\textcolor{mpcblue800}{Obs. Encoder}};
\node[lat] (zt)    at (-4.4, -2.8) {\textcolor{mpcteal800}{$z_t$}};
\draw[->, line width=0.5pt, color=mpcgray600] (St)    -- (Eleft);
\draw[->, line width=0.5pt, color=mpcblue600] (Eleft) -- (zt);

\node[obs] (amask) at (-1.2, -6.4) {\textcolor{mpcgray800}{$a_{t:t+k-1}$}};
\node[enc] (aenc)  at (-1.2, -2.8) {\textcolor{mpcblue800}{Act. Encoder}};
\node[lat] (lt)    at (-1.2, -1.0) {\textcolor{mpcteal800}{$l_t$}};
\draw[->, line width=0.5pt, color=mpcgray600] (amask) -- (aenc);
\draw[->, line width=0.5pt, color=mpcteal600] (zt)    -- (aenc);   %
\draw[->, line width=0.5pt, color=mpcblue600] (aenc)  -- (lt);

\node[pred] (obspred) at (1.6,  0.0) {\textcolor{mpcblue800}{Obs. Predictor}};
\node[pred] (actdec)  at (1.6,  1.6) {\textcolor{mpcblue800}{Act. Decoder}};

\draw[->, line width=0.5pt, color=mpcteal600]
  (zt.north) -- (-4.4, 1.8) -- (actdec.west |- 0,1.8);
\draw[->, line width=0.5pt, color=mpcteal600]
  (-4.4, 0.2) -- (obspred.west |- 0,0.2);
\node[junction] at (-4.4, 0.2) {};

\draw[->, line width=0.5pt, color=mpcteal600, bridge]
  (lt.north) -- (-1.2, 1.4) -- (actdec.west |- 0,1.4);
\draw[->, line width=0.5pt, color=mpcteal600]
  (-1.2, -0.2) -- (obspred.west |- 0,-0.2);
\node[junction] at (-1.2, -0.2) {};

\node[obs]  (Stk)    at (5.2, -6.4) {\textcolor{mpcgray800}{$s_{t+k}$}};
\node[enc]  (Eright) at (5.2, -4.6) {\textcolor{mpcblue800}{Obs. Encoder}};
\node[lat]  (zt1)    at (5.2, -2.8) {\textcolor{mpcteal800}{$z_{t+k}$}};
\node[zhat] (zhat)   at (5.2,  0.0) {\textcolor{mpcblue800}{$\hat{z}_{t+k}$}};
\node[mse]  (mseobs) at (5.2, -1.4) {\textcolor{mpcred800}{\small MSE}};

\draw[->, line width=0.5pt, color=mpcgray600] (Stk)    -- (Eright);
\draw[->, line width=0.5pt, color=mpcblue600] (Eright) -- (zt1);
\draw[->, line width=0.5pt, color=mpcblue600] (obspred) -- (zhat);
\draw[->, line width=0.5pt, dashed, color=mpcblue600] (zhat) -- (mseobs);
\draw[->, line width=0.5pt, dashed, color=mpcteal600] (zt1)  -- (mseobs);

\node[obs]  (atgt)   at (7.8, -6.4) {\textcolor{mpcgray800}{$a_{t:t+k-1}$}};
\node[zhat] (ahat)   at (7.8,  0.0) {\textcolor{mpcblue800}{$\hat{a}_{t:t+k-1}$}};
\node[mse]  (mseact) at (7.8, -2.8) {\textcolor{mpcred800}{\small MSE}};

\draw[->, line width=0.5pt, color=mpcblue600] (actdec.east) -- (7.8, 1.6) -- (ahat);
\draw[->, line width=0.5pt, dashed, color=mpcblue600] (ahat) -- (mseact);
\draw[->, line width=0.5pt, dashed, color=mpcgray600] (atgt) -- (mseact);

\node[anchor=west, font=\small\bfseries, color=mpcgray800] at (-5.55, 3.15) {(a) Training};
\begin{scope}[on background layer]
\node[fit=(current bounding box), inner sep=6pt, rounded corners=8pt,
      fill=panelA, draw=panelAborder, line width=0.4pt] {};
\end{scope}

\end{tikzpicture}
}
\phantomsubcaption\label{fig:skill-jepa-train}
\end{minipage}\hfill
\begin{minipage}[t]{0.495\textwidth}
\centering
\resizebox{\linewidth}{!}{%
\begin{tikzpicture}[
  >=Stealth,
  x=0.7cm, y=0.8cm,
  font=\small,
  obs/.style={
    rectangle, rounded corners=5pt,
    fill=mpcgray50, draw=mpcgray600, line width=0.4pt,
    minimum width=1.6cm, minimum height=0.7cm, inner sep=4pt
  },
  enc/.style={
    rectangle, rounded corners=5pt,
    fill=mpcblue50, draw=mpcblue600, line width=0.4pt,
    minimum width=1.6cm, minimum height=0.7cm, inner sep=4pt
  },
  lat/.style={
    rectangle, rounded corners=5pt,
    fill=mpcteal50, draw=mpcteal600, line width=0.4pt,
    minimum width=1.6cm, minimum height=0.7cm, inner sep=4pt
  },
  pred/.style={
    rectangle, rounded corners=5pt,
    fill=mpcblue50, draw=mpcblue600, line width=0.4pt,
    minimum width=2.2cm, minimum height=0.7cm, inner sep=4pt
  },
  zhat/.style={
    rectangle, rounded corners=5pt,
    fill=mpcblue50, draw=mpcblue600, line width=0.4pt,
    dashed,
    minimum width=1.7cm, minimum height=0.7cm, inner sep=4pt
  },
  cem/.style={
    rectangle, rounded corners=5pt,
    fill=mpcamber50, draw=mpcamber600, line width=0.4pt,
    minimum width=1.6cm, minimum height=0.7cm, inner sep=4pt
  },
  mse/.style={
    rectangle, rounded corners=4pt,
    fill=mpcred50, draw=mpcred600, line width=0.4pt,
    minimum width=1.1cm, minimum height=0.55cm, inner sep=3pt
  },
  junction/.style={circle, fill=mpcteal600, inner sep=0pt, minimum size=2.4pt},
  bridge/.style={preaction={draw=panelB, line width=2.4pt}}
]

\node[obs] (St)    at (-4.4, -6.4) {\textcolor{mpcgray800}{$s_t$}};
\node[enc] (Eleft) at (-4.4, -4.6) {\textcolor{mpcblue800}{Obs. Encoder}};
\node[lat] (zt)    at (-4.4, -2.8) {\textcolor{mpcteal800}{$z_t$}};
\draw[->, line width=0.5pt, color=mpcgray600] (St)    -- (Eleft);
\draw[->, line width=0.5pt, color=mpcblue600] (Eleft) -- (zt);

\node[obs] (prior) at (-1.2, -6.4) {\textcolor{mpcgray800}{$p(l)=\mathcal{N}(0,I)$}};
\node[cem] (cem)   at (-1.2, -2.8) {\textcolor{mpcamber800}{CEM Sampler}};
\node[lat] (lt)    at (-1.2, -1.4) {\textcolor{mpcteal800}{$l_{1:M}^{(1:K)}$}};
\draw[->, line width=0.5pt, color=mpcgray600] (prior) --
  node[right, font=\scriptsize, color=mpcgray600] {$\pi_0 = p$} (cem);
\draw[->, line width=0.5pt, color=mpcamber600] (cem) -- (lt);

\node[pred] (obspred) at (1.6,  0.0) {\textcolor{mpcblue800}{Obs. Predictor}};
\node[pred] (actdec)  at (1.6,  2.2) {\textcolor{mpcblue800}{Act. Decoder}};

\draw[->, line width=0.5pt, color=mpcteal600]
  (zt.north) -- (-4.4, 2.4) -- (actdec.west |- 0,2.4);
\draw[->, line width=0.5pt, color=mpcteal600]
  (-4.4, 0.2) -- (obspred.west |- 0,0.2);
\node[junction] at (-4.4, 0.2) {};

\draw[line width=0.5pt, color=mpcteal600] (lt.north) -- (-1.2, -0.2);
\draw[->, line width=0.5pt, color=mpcteal600]
  (-1.2, -0.2) -- (obspred.west |- 0,-0.2);
\draw[->, line width=0.5pt, dashed, color=mpcamber600, bridge]
  (-1.2, -0.2) --
  node[left, font=\scriptsize, color=mpcamber800, pos=0.7] {$l^{\star}_{1:M}$} (-1.2, 2.0) --
  (actdec.west |- 0,2.0);
\node[junction] at (-1.2, -0.2) {};

\node[obs]  (Sg)     at (5.2, -6.4) {\textcolor{mpcgray800}{$s_g$}};
\node[enc]  (Eright) at (5.2, -4.6) {\textcolor{mpcblue800}{Obs. Encoder}};
\node[lat]  (zg)     at (5.2, -2.8) {\textcolor{mpcteal800}{$z_g$}};
\node[zhat] (zhat)   at (5.2,  0.0) {\textcolor{mpcblue800}{$\hat{z}_{t+H}^{(1:K)}$}};
\node[mse]  (cost)   at (5.2, -1.4) {\textcolor{mpcred800}{\small cost}};

\draw[->, line width=0.5pt, color=mpcgray600] (Sg)     -- (Eright);
\draw[->, line width=0.5pt, color=mpcblue600] (Eright) -- (zg);
\draw[->, line width=0.5pt, color=mpcblue600] (obspred) -- (zhat);
\draw[->, line width=0.5pt, dashed, color=mpcblue600] (zhat) -- (cost);

\draw[->, line width=0.5pt, color=mpcblue600]
  (zhat.north) -- (5.2, 1.0) --
  node[above, font=\scriptsize, color=mpcblue800, pos=0.5] {rollout}
  (2.6, 1.0) -- (2.6, 0 |- obspred.north);
\draw[->, line width=0.5pt, dashed, color=mpcteal600] (zg)   -- (cost);

\draw[->, line width=0.5pt, color=mpcred600]
  (cost.west) -- (1.6, -1.4) --
  node[right, font=\scriptsize, color=mpcred800, align=left] {refit\\elites}
  (1.6, -2.8) -- (cem.east);

\node[zhat] (ahat) at (7.8,  0.0) {\textcolor{mpcblue800}{$\hat{a}_{t:t+H-1}$}};
\node[obs]  (env)  at (7.8, -6.4) {\textcolor{mpcgray800}{Env.}};
\draw[->, line width=0.5pt, color=mpcblue600] (actdec.east) -- (7.8, 2.2) -- (ahat);
\draw[->, line width=0.5pt, color=mpcgray600] (ahat) --
  node[above, sloped, font=\scriptsize, color=mpcgray600, pos=0.5] {execute $C$ steps} (env);

\node[anchor=west, font=\small\bfseries, color=mpcgray800] at (-5.55, 3.15) {(b) Inference: CEM planning};
\begin{scope}[on background layer]
\node[fit=(current bounding box), inner sep=6pt, rounded corners=8pt,
      fill=panelB, draw=panelBborder, line width=0.4pt] {};
\end{scope}

\end{tikzpicture}
}
\phantomsubcaption\label{fig:skill-jepa-infer}
\end{minipage}
\caption{Overview. (a) Training: a conditional VAE encodes action chunks into skills
  $l_t$ given the latent observation $z_t$, and the world model is trained to predict
  the next latent observation and to reconstruct the actions. (b) Inference: the action
  encoder is replaced by a sampling-based planner initialized at the skill prior,
  $\pi_0 = p(l)$, which searches over skill sequences with the predictor and decodes
  the selected plan into actions (Sec.~\ref{sec:planners}).}
\label{fig:skill-jepa}
\end{figure*}
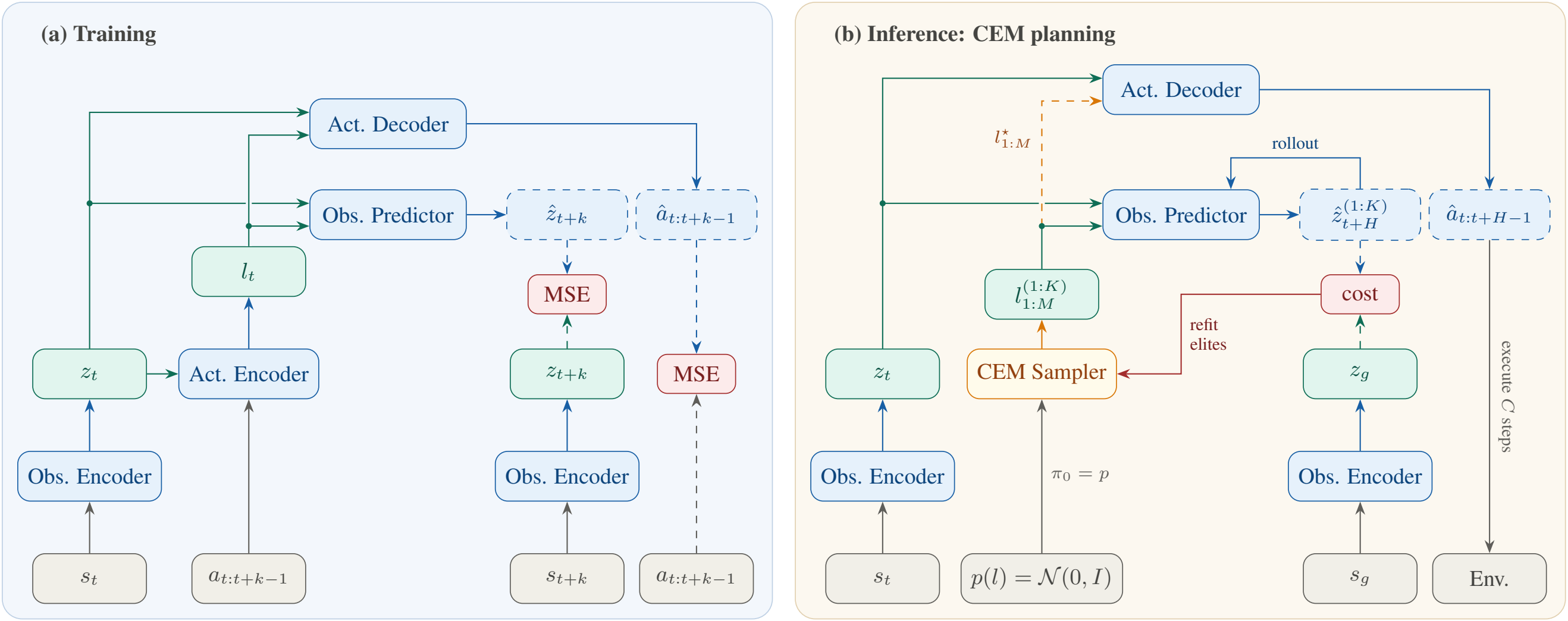

We take a different route: we change neither the planner nor the data, but the action space that connects them. Recent JEPA world models learn their state representation end-to-end~\cite{sobal2025learningrewardfreeofflinedata, maes_lelidec2026lewm, ivashkov2026sensorimotorworldmodelsperception}. We let them learn their action space as well, so that the planner's samples fall where the model was trained. Our method, Skill-JEPA (Figure~\ref{fig:skill-jepa}), adds a variational autoencoder~\cite{kingma2022autoencodingvariationalbayes} to the world model. Conditioned on the current state, it encodes short action sequences into latent codes with a standard Gaussian prior, and decodes them back into actions. We call these codes \emph{skills}, following \citet{pertsch2020acceleratingreinforcementlearninglearned}. The world model is trained to predict the effect of skills rather than raw actions, and the planner searches over skills, starting from that same Gaussian. Because the decoder sees the state, the same sample becomes a different action sequence in each state, one that resembles what the data shows there. The dependence on the state is thus handled by the decoder, and a single Gaussian can serve the planner everywhere.

Skill-JEPA is deliberately simple. Its only addition is the autoencoder, trained end-to-end with the world model, and the planner is used as is. In effect, the decoder teaches the planner what realistic action sequences look like, but without training a policy for any task: the knowledge is built into the action space itself. Skill-JEPA is also not hierarchical. Skills do not make the planner look farther ahead, since our raw-action baselines plan over action chunks of the same length, and there are no subgoals or higher-level models. Only the representation of each chunk changes, which makes the method complementary to the hierarchical approaches above.
Nor is the building block new: latent codes for action sequences are widely used in imitation and reinforcement learning (Section~\ref{sec:related-work}).

What we add is a precise role for the prior. We make it the distribution that a sampling-based planner starts from, so that the term that regularizes the autoencoder also limits how much worse the model can be on the planner's initial candidates than on its training data (Section~\ref{sec:prior-mismatch}). We note that the prior is Gaussian because that is where such planners start, not because the idea requires it: the analysis holds for any initial distribution, and a planner that starts elsewhere would call for a different prior.

In summary, our contributions are:
\begin{itemize}
  \item \textbf{Method.} Skill-JEPA aligns the action space of a JEPA world model with its planner through a state-conditioned action autoencoder whose prior is the planner's initial sampling distribution. The planner is left unchanged.
  \item \textbf{Analysis.} We bound the model's excess prediction error on the planner's initial candidates through the autoencoder's KL term, and show with controlled experiments that the gains vanish without this term.
  \item \textbf{Evaluation.} On three JEPA world models, five tasks and four planners, the gains grow with goal distance and hold for every planner. Average success on goals 80 steps away rises from 42\% to 62\%, and on AntMaze goals two maze cells away, from 0.4\% to 38\%.
\end{itemize}

\section{Related work}
\label{sec:related-work}

\paragraph{JEPA world models.}
World models learn how an environment responds to actions~\cite{ha2018worldmodels, hafner2020dream}. JEPA world models~\cite{lecun-autonomous-machine-intelligence-2022} learn this in a latent space, and need a mechanism that keeps the space from collapsing. Some rely on a frozen pretrained encoder~\cite{zhou2025dinowmworldmodelspretrained, assran2025vjepa2selfsupervisedvideo}. Others train the encoder end-to-end with a regularizer: variance regularization with inverse dynamics~\cite{sobal2025learningrewardfreeofflinedata}, inverse dynamics alone~\cite{ivashkov2026sensorimotorworldmodelsperception}, or a term that pushes the embeddings toward an isotropic Gaussian~\cite{balestriero2025lejepaprovablescalableselfsupervised, maes_lelidec2026lewm}. We apply Skill-JEPA to these three end-to-end models. %
They are used with sampling-based planners~\cite{Rubinstein1999, williams2017informationtheoreticmodelpredictive, pinneri2020sampleefficientcrossentropymethodrealtime, howell2022predictivesamplingrealtimebehaviour}, whose role is studied by \citet{terver2025jepawms}; we leave these planners unchanged.

\paragraph{Planning away from the training data.}
A learned model is reliable only near its training data, and optimizers exploit its errors elsewhere~\cite{yu2020mopo, janner2022diffuser}. For latent world models, accuracy on the training data does not predict planning success, whereas accuracy on the states that the planner reaches does~\cite{you2026control}. Existing remedies adapt the model, by fine-tuning it on planner-generated or adversarially perturbed trajectories~\cite{agrawal2025closing}, or keep the optimizer close to the data, by penalizing uncertain predictions~\cite{yu2020mopo} or by drawing candidates from a learned policy~\cite{argenson2021mbop, hansen2024tdmpc2, chahe2026pijepa}. In concurrent work, LeWAM~\cite{hegde2026lewam} searches over the noise of a generative policy trained jointly with a JEPA world model; its predictor takes the decoded actions as input. Skill-JEPA needs neither a separate policy nor a fine-tuning stage. It changes the action space so that an unmodified planner already samples where the model was trained, and its predictor is trained directly on the variable that the planner searches.

\paragraph{Latent action spaces.}
Encoding short action sequences into latent codes with a variational autoencoder is a common tool: to learn policies from demonstrations~\cite{zhao2023learningfinegrainedbimanualmanipulation}, to extract reusable skills from offline data~\cite{pertsch2020acceleratingreinforcementlearninglearned, ajay2021opal}, and to keep offline policies within the support of the data~\cite{zhou2020plas}. Such codes have also been used for planning. \citet{mishra2017prediction} learn a prior over action segments so that trajectory optimization stays in the regime where the model is valid, which is also our motivation; their dynamics model takes the decoded actions as input, and plans are found by gradient-based optimization. Other methods learn dynamics over skills~\cite{shi2022skimo} or search over discrete codes~\cite{jiang2023tap}, but rely on rewards and on a learned prior or policy over codes. Skill-JEPA is reward-free and keeps the prior fixed. Since the encoder and decoder absorb the dependence on the state, the prior can be exactly the Gaussian that sampling-based planners draw from, and the model is trained on those same codes.

\paragraph{Long-horizon planning.}
Planning success drops as goals move farther away, and several recent methods change how far ahead the model or the planner looks. Hierarchical world models plan at several time scales, with upper levels that set subgoals for lower ones~\cite{zhang2026hierarchicalplanninglatentworld}, including concurrent work~\cite{zhang2026hjepaendtoendlearninghierarchical}. Others learn a model that proposes subgoals~\cite{masip2026ffjepa}, or train a single model to predict across jumps of variable length~\cite{du2026vlwm}. Skill-JEPA is not hierarchical and does not change the time scale: it uses one model and one planner, and its skills span as many steps as the action chunks of our raw-action baselines. It targets a different cause of failure, the mismatch between what the planner samples and what the model was trained on, and is complementary to these methods.

\makeatletter
\newcommand{\LineComment}[1]{\Statex \hskip\dimexpr\ALG@tlm-\algorithmicindent\relax \(\triangleright\) \textit{#1}}
\makeatother

\section{Method}
\label{sec:methods}

Skill-JEPA changes a JEPA world model in one place: the predictor is conditioned on a learned code for the next actions, a \emph{skill}, instead of on the actions themselves. In this section, we describe the model and its training objective, how a standard planner searches over skills, and why the prior over skills should be the distribution that the planner samples from. Figure~\ref{fig:skill-jepa} gives an overview.

\subsection{A world model over skills}
\label{sec:skill-vae}

We consider a reward-free Markov decision process with states $s \in \mathcal{S}$ (images, in our experiments), actions $a \in \mathcal{A}$, and transition dynamics $s_{t+1} \sim \mathcal{T}(\cdot \mid s_t, a_t)$. We are given an offline dataset $\mathcal{D}$ of trajectories $(s_0, a_0, \dots, a_{T-1}, s_T)$, without rewards. At test time, the agent receives a goal state $s_g$ and must reach it.

A JEPA world model \citep{lecun-autonomous-machine-intelligence-2022} has an encoder, which maps a state to a latent observation, and a predictor, which predicts the latent observation $k$ steps ahead:
\begin{equation}
  z_t = f_\theta(s_t), \qquad \hat z_{t+k} = g_\theta(z_t, c_t).
  \label{eq:jepa}
\end{equation}
The input $c_t$ tells the predictor what the agent does during these $k$ steps. In a standard model it is the chunk of raw actions, $c_t = a_{t:t+k-1}$ \citep{zhao2023learningfinegrainedbimanualmanipulation}. In Skill-JEPA it is a learned code for that chunk, a skill, $c_t = l_t$ \citep{pertsch2020acceleratingreinforcementlearninglearned}. Both models thus make one decision every $k$ steps; they differ only in how that decision is represented.

We learn skills with a conditional variational autoencoder (VAE) \citep{kingma2022autoencodingvariationalbayes, sohn2015cvae}. Writing $a$ for a chunk $a_{t:t+k-1}$ and $z$ for the latent observation at which it starts, the encoder maps the chunk to a Gaussian over skills $l \in \mathbb{R}^d$,
\begin{equation}
  q_\theta(l \mid a, z) =
  \mathcal{N}\!\left(\mu_\theta(a, z),\, \operatorname{diag}\sigma^2_\theta(a, z)\right),
  \label{eq:skill-vae}
\end{equation}
and the decoder $\hat a_\theta(l, z)$ maps a skill back to a chunk. Both are conditioned on $z$, so the same skill can stand for different actions in different states. The prior over skills, in contrast, is the same in every state, $p(l) = \mathcal{N}(0, I)$. A conditional VAE would normally learn a state-dependent prior $p(l \mid z)$ \citep{sohn2015cvae}. Like ACT \citep{zhao2023learningfinegrainedbimanualmanipulation}, we keep it fixed, because a fixed Gaussian is exactly what a planner samples from.%

\paragraph{Training objective.}
All modules are trained jointly, without stopping gradients, %
with three terms defined on the data,

\begin{subequations}
\label{eq:losses}
\begin{align}
  \mathcal{L}_{\text{pred}} &\coloneqq
    \mathbb{E}\!\left[\bigl\| g_\theta(z_t, l_t) - z_{t+k} \bigr\|_2^2\right], \label{eq:loss-pred} \\
  \mathcal{L}_{\text{rec}} &\coloneqq
    \mathbb{E}\!\left[\bigl\| \hat a_\theta(l_t, z_t) - a_{t:t+k-1} \bigr\|_2^2\right], \label{eq:loss-rec} \\
  \mathcal{L}_{\text{KL}} &\coloneqq
    \mathbb{E}\!\left[\mathrm{KL}\!\left(q_\theta(l \mid a_{t:t+k-1}, z_t) \,\|\, p(l)\right)\right], \label{eq:loss-kl}
\end{align}
\end{subequations}
where $l_t \sim q_\theta(\cdot \mid a_{t:t+k-1}, z_t)$, and a fourth term $\mathcal{L}_{\text{reg}}$ that keeps the latent observations from collapsing:
\begin{equation}
  \mathcal{L} \coloneqq \mathcal{L}_{\text{pred}} + \mathcal{L}_{\text{rec}}
  + \beta\, \mathcal{L}_{\text{KL}} + \lambda\, \mathcal{L}_{\text{reg}}.
  \label{eq:objective}
\end{equation}
The prediction term trains the model to predict the effect of skills. The reconstruction term makes skills decodable into actions. The KL term pulls the skills produced by the encoder toward the prior; together with the reconstruction term, it forms the $\beta$-VAE objective \citep{higgins2017betavae} (Appendix~\ref{app:skill-objective}). The regularizer $\mathcal{L}_{\text{reg}}$ is the one of the base world model (Section~\ref{sec:results}). A raw-action model drops $\mathcal{L}_{\text{rec}}$ and $\mathcal{L}_{\text{KL}}$, and its predictor takes the action chunk as input.

\subsection{Planning over skills}
\label{sec:planners}

We plan with standard sampling-based optimizers, which we do not modify (Figure~\ref{fig:skill-jepa}b). To reach a goal $s_g$ from a state $s_t$ with a horizon of $H$ steps, the planner searches over sequences of $M = H/k$ skills -- where $k$ is the length of the action chunk. It maintains a Gaussian over such sequences, $\pi_\phi(l_{1:M}) = \prod_{j=1}^{M}\mathcal{N}(\mu_j, \operatorname{diag}\sigma_j^2)$, and repeats three steps for $I$ iterations. First, it samples $K$ candidate sequences from $\pi_\phi$. Second, it scores each candidate: starting from $z_t = f_\theta(s_t)$, it applies the predictor once per skill and measures how far the last prediction is from the goal, $J = \lVert \hat z_{t+H} - f_\theta(s_g) \rVert_2^2$. Third, it refits $\pi_\phi$ to the $E$ candidates with the lowest cost. The final mean $\mu_{1:M}$ is the plan (Algorithm~\ref{alg:sampling-planner} in Appendix~\ref{app:planning-algos}).

The plan is a sequence of skills, $l^\star_{1:M} = \mu_{1:M}$, which we decode into actions one skill at a time. The decoder needs the latent observation at which each skill starts. The first skill starts at the current state, so its actions are $\hat a_\theta(l^\star_1, z_t)$. The second starts $k$ steps later, at a state that the agent has not reached yet. We therefore use the model's prediction, $\hat z_{t+k} = g_\theta(z_t, l^\star_1)$, and decode $\hat a_\theta(l^\star_2, \hat z_{t+k})$. The remaining skills are decoded in the same way. The agent executes the first $C$ actions and replans from the state it reaches.

At every replanning step we initialize the planner at the skill prior, $\pi_0 = p(l) = \mathcal{N}(0, I)$, so its first candidates are samples from the prior and nothing is carried over between replanning steps. Nothing else about the skill model has to be given to the planner, and the planner itself is unchanged: only the meaning of the variable it searches is different. With raw actions, the same planner searches over standardized action chunks.

We use CEM \citep{Rubinstein1999} as our default planner and additionally evaluate iCEM \citep{pinneri2020sampleefficientcrossentropymethodrealtime}, MPPI \citep{williams2017informationtheoreticmodelpredictive} and Predictive Sampling (PS) \citep{howell2022predictivesamplingrealtimebehaviour}. All four follow the template above \citep{wagener2019onlinelearningapproachmodel} with different rules for weighting candidates and refitting (Appendix~\ref{app:planning-algos}). PS is the simplest: in a single iteration, it keeps the best candidate drawn from the prior, which is random shooting.

\subsection{Why the prior should match the planner}
\label{sec:prior-mismatch}

The predictor is trained only on skills produced by the encoder. At a state $s$ with $z = f_\theta(s)$, these skills follow the \emph{aggregate posterior}
\begin{equation}
  q_{\text{agg}}(l \mid z) = \mathbb{E}_{a \sim p_{\text{data}}(a \mid s)}\!\left[q_\theta(l \mid a, z)\right],
  \label{eq:agg-posterior}
\end{equation}
where $p_{\text{data}}(a \mid s)$ is the distribution of the action chunks that follow $s$ in the dataset. A planner knows nothing of this distribution. If it proposes skills to which $q_{\text{agg}}(\cdot \mid z)$ gives little mass, the predictor is queried away from its training data, and its predictions, and with them the costs used to rank candidates, become unreliable. We call this the \emph{sampling-prior mismatch}.

Formally, let $e(s, l)$ be the error that the predictor makes when skill $l$ is used at state $s$, measured against the state that is actually reached by executing the decoded chunk:
\begin{equation}
  e(s, l) = \mathbb{E}_{s'}\bigl\|\, g_\theta(z, l) - f_\theta(s') \,\bigr\|_2,
  \label{eq:pred-error}
\end{equation}
where $s' \sim \mathcal{T}^k\!\left(\cdot \mid s,\, \hat a_\theta(l, z)\right)$ and $\mathcal{T}^k$ applies the dynamics for $k$ steps. This error bounds how far the predicted distance of a candidate to the goal $z_g$ can be from its true distance. By the triangle inequality,
\begin{equation}
  \bigl|\, \|g_\theta(z, l) - z_g\|_2 - \mathbb{E}_{s'}\|f_\theta(s') - z_g\|_2 \,\bigr| \le e(s, l),
\end{equation}
so a smaller $e(s,l)$ means the planner ranks candidates more reliably. %
Since encoder and predictor outputs are layer-normalized, latents have bounded norm, so $e(s,l) \le B$ for some constant $B$. (Appendix~\ref{app:proofs}).

\begin{prop}
For every state $s$, with $z = f_\theta(s)$, and every initial distribution $\pi_0$,
\begin{equation}
\begin{split}
  \mathbb{E}_{l \sim \pi_0}\!\left[e(s, l)\right]
  \le\; &\mathbb{E}_{l \sim q_{\text{agg}}(\cdot \mid z)}\!\left[e(s, l)\right] \\
  &+ B \sqrt{\tfrac{1}{2}\,\mathrm{KL}\!\left(q_{\text{agg}}(\cdot \mid z)\,\|\,\pi_0\right)}.
\end{split}
\label{eq:prop1}
\end{equation}
\end{prop}

In other words, on the candidates that the planner samples first, the model is at most as wrong as on the skills it was trained on, plus a penalty that grows with the mismatch between the two distributions and vanishes when they coincide. The first term averaged over states is equivalent to the prediction loss Eq.~\eqref{eq:loss-pred} and is minimized through training. Setting $\pi_0 = p$ similarly puts the penalty under the control of training: averaged over states, it is bounded through the KL term of Eq.~\eqref{eq:loss-kl} (Corollary~1, Appendix~\ref{app:proofs}).
That term equals the mismatch plus the information that skills carry about the actions, which cannot be negative \citep{hoffman2016elbo}. The term that regularizes the autoencoder is therefore also the one that limits the mismatch.

We note that this result concerns the planner's first candidates and a single prediction step.
It assumes nothing about $\pi_0$ or about how the planner refines its search, so it applies to the first candidates of any planner that starts from the prior, sampling-based or not.
As the planner refits its distribution, the bound no longer applies directly (Section~\ref{sec:discussions}). Nor is the bound meant to be tight: its role is to identify the quantity that training should control. Section~\ref{sec:results} tests this role by removing the KL term, and by matching the planner to the skills only through their mean and covariance.

\section{Results}
\label{sec:results}

Our experiments address three questions. Does the same planner reach farther goals when it searches over skills instead of raw actions? If so, is it because the skills are aligned with the distribution that the planner samples from, as Section~\ref{sec:prior-mismatch} argues? And do the gains depend on the planner?

\begin{figure}[t]
  \centering
  \includegraphics[width=\linewidth]{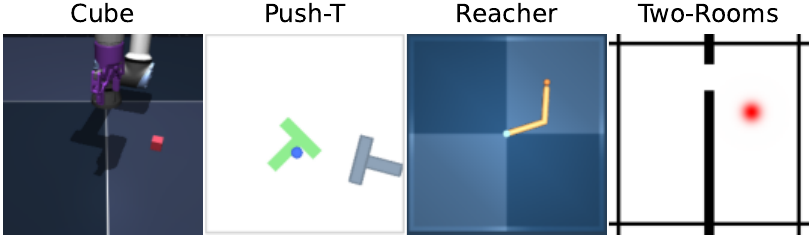}
  \caption{The four standard tasks. AntMaze is shown in Figure~\ref{fig:teaser}.}
  \label{fig:task_frames}
\end{figure}

\begin{figure*}[t]
  \centering
  \includegraphics[width=\linewidth]{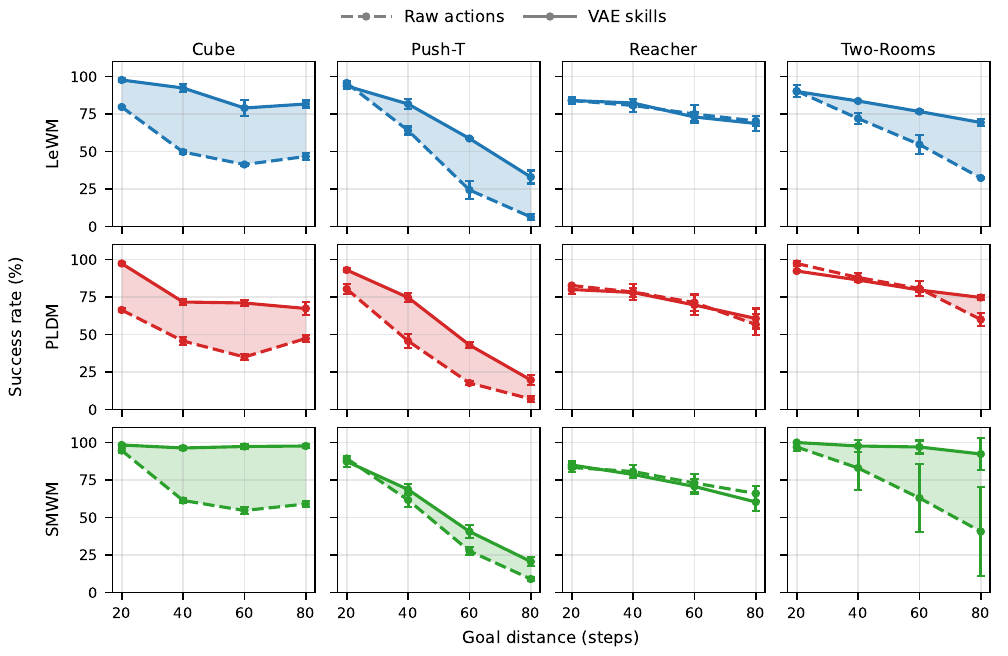}
  \caption{\textbf{Success falls more slowly with goal distance when planning over skills.} Success rate on the four standard tasks (columns) for three JEPA world models (rows), with CEM searching over raw actions (dashed) or over VAE skills (solid); the shaded area is the gap between them. Mean and standard deviation over $3$ training seeds, with $100$ episodes each.}
  \label{fig:all/success_rate_vs_distance}
\end{figure*}

\subsection{Experimental setup}
We evaluate on five tasks, each with an offline dataset of trajectories. Four are standard benchmarks for JEPA world models (Figure~\ref{fig:task_frames}): navigation in Two-Rooms \citep{sobal2025learningrewardfreeofflinedata}, manipulation in Push-T \citep{chi2024diffusionpolicyvisuomotorpolicy} and OGBench-Cube \citep{park2025ogbenchbenchmarkingofflinegoalconditioned}, and control of a two-link arm in Reacher \citep{tassa2018deepmindcontrolsuite}. For these we use the datasets and the evaluation protocol of \citet{maes_lelidec2026lewm}, which follows \citet{zhou2025dinowmworldmodelspretrained} for Push-T and Reacher, so that our baselines follow an established setup. The fifth is OGBench-AntMaze \citep{park2025ogbenchbenchmarkingofflinegoalconditioned}, with the dataset of \citet{zhang2026hjepaendtoendlearninghierarchical}: a quadruped with eight actuated joints must cross sections of a maze (Figure~\ref{fig:teaser}).

We apply Skill-JEPA to three JEPA world models that train their encoder end-to-end and differ in how they prevent collapse. LeWorldModel (LeWM) \citep{maes_lelidec2026lewm} uses SIGReg \citep{balestriero2025lejepaprovablescalableselfsupervised}, Planning with a Latent Dynamics Model (PLDM) \citep{sobal2025learningrewardfreeofflinedata} combines variance-covariance regularization with a temporal smoothness term, and the Sensori-Motor World Model (SMWM) \citep{ivashkov2026sensorimotorworldmodelsperception} relies on inverse dynamics. This tests whether the effect of skills is tied to one regularizer.

Each skill model is compared with the same world model trained on raw actions. The two share the encoder and predictor architectures, the regularizer, the data and the chunk length, $k = 4$: the baseline's predictor takes the chunk of four actions, standardized to zero mean and unit variance, whereas ours takes a skill. They are planned with the same planner and settings, which searches over standardized chunks in one case and over skills in the other. Both models thus make one decision every four steps, and the comparison isolates how that decision is represented. Figures label the two as \emph{raw actions} and \emph{VAE skills}.

The agent is given a goal image and plans with CEM \citep{Rubinstein1999} unless stated otherwise. Since prediction errors compound along a rollout, we expect the action space to matter most for distant goals, and we vary the goal distance. On the four standard tasks, goals are $20$, $40$, $60$ and $80$ steps away; the shortest distance is close to the $25$ steps of the standard protocol \citep{maes_lelidec2026lewm}. On AntMaze, following \citet{zhang2026hjepaendtoendlearninghierarchical}, goals are one to four maze cells away. We report success rates over $100$ episodes per model, task, distance and seed, with three training seeds; error bars show the standard deviation across seeds. Success criteria and planner settings are given in Appendix~\ref{app:eval-details}.

\subsection{Skills reach farther goals}%
Figure~\ref{fig:all/success_rate_vs_distance} compares the two action spaces as goals move away. With raw actions, success falls with distance for every world model: averaged over tasks and models, from $87\%$ at $20$ steps to $42\%$ at $80$ steps. With skills it falls more slowly, from $92\%$ to $62\%$. The gap (shaded) therefore grows with distance, from $5$ percentage points at $20$ steps to $20$ at $80$ steps, where skills are better in $10$ of the $12$ combinations of task and world model (Appendix~\ref{app:results}). In addition, all three world models benefit. At $80$ steps, the average gain is $13$ points for PLDM and about $24$ for LeWM and SMWM. The exception is Reacher, where raw actions degrade little with distance and skills make no difference.

Our most demanding task is the AntMaze. In this task, each action chunk has $32$ dimensions, the legs must move in coordination for the robot to advance, and goals are several maze cells away (Figure~\ref{fig:teaser}). Over raw actions, the planner reaches only the nearest goals: averaged over world models, success is $49\%$ at one cell, $0.4\%$ at two, and zero beyond. Over skills, the same planner reaches $83\%$ at one cell, $38\%$ at two and $21\%$ at three, and still succeeds occasionally at four ($6\%$).
We stress that nothing else changed: the data, the architectures, the planner and its settings are the same in both cases. The action space alone thus decides whether goals beyond the nearest cell are within the planner's reach.%

\begin{figure}[t]
  \centering
  \includegraphics[width=\linewidth]{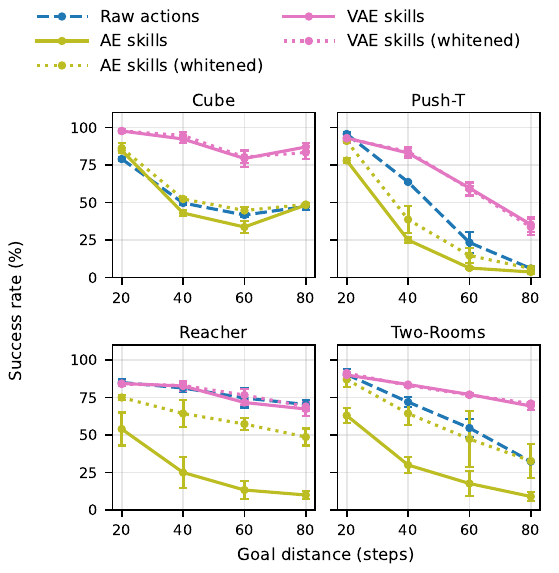}
  \caption{\textbf{The gains require the KL term.} Success rate of LeWM with CEM over raw actions, over skills learned without the KL term (AE skills), and over VAE skills. Dotted lines start CEM from a Gaussian fitted to the skills (whitened): this brings AE skills at most to about the level of raw actions, and leaves VAE skills unchanged.}
  \label{fig:lewm_latent_actions/success_rate_vs_distance}
\end{figure}

\begin{figure*}[t]
  \centering
  \begin{subfigure}[t]{0.49\linewidth}
    \centering
    \includegraphics[width=\linewidth]{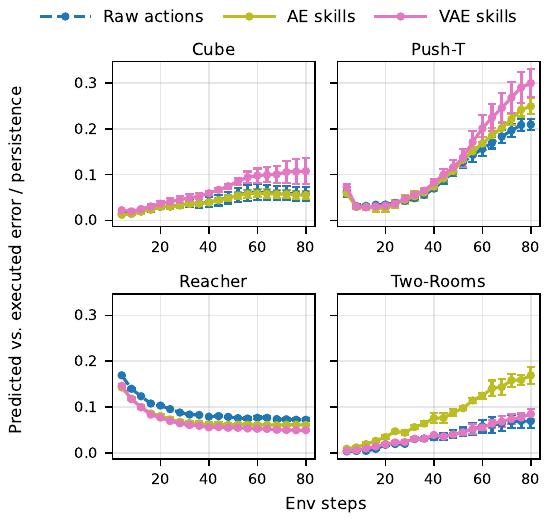}
    \caption{Action sequences from the dataset.}
    \label{fig:lewm_latent_actions/control_plan_execution_by_step}
  \end{subfigure}\hfill
  \begin{subfigure}[t]{0.49\linewidth}
    \centering
    \includegraphics[width=\linewidth]{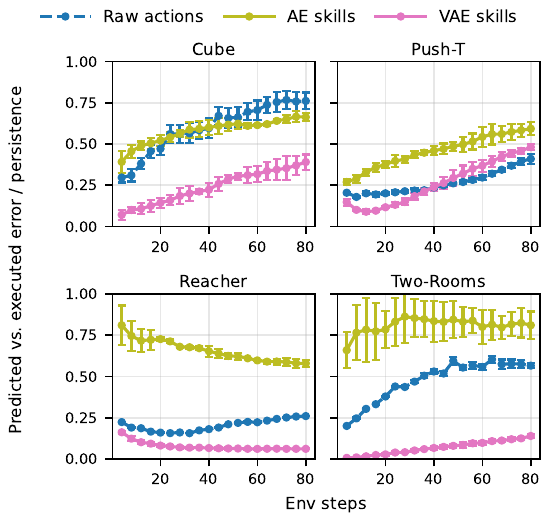}
    \caption{Candidates sampled from $\pi_0$ by the planner.}
    \label{fig:lewm_latent_actions/iteration0_plan_execution_by_step}
  \end{subfigure}
  \caption{\textbf{Skills make the model accurate where the planner samples, not on the data.} Prediction error of LeWM along rollouts of up to $80$ steps: distance between the predicted latent observation and the encoding of the state actually reached, relative to a persistence baseline that predicts no change (lower is better). (a) On the dataset, the three models are comparable. (b) On the planner's initial candidates, the VAE model has the lowest error, except on Push-T beyond $40$ steps. Note the different vertical scales.}
  \label{fig:pred-error-dataset-vs-planner}
\end{figure*}

\subsection{Planner alignment drives the gains}
Section~\ref{sec:prior-mismatch} attributes the benefit of skills to the KL term, which aligns the skills seen in training with the distribution $\pi_0$ that the planner starts from. But a skill model differs from a raw-action model in other ways: it has an extra encoder and decoder, whose losses also shape the observation encoder through $z_t$. We separate these factors with three experiments on LeWM and the four standard tasks.

\begin{table}[t]
\caption{Mismatch between the skills of LeWM and the planner's initial distribution: $\mathrm{KL}(\hat q \,\|\, \mathcal{N}(0, I))$ in nats per skill, where $\hat q$ is a diagonal Gaussian fitted to the skills of the training set (sampled, for VAE skills). Mean $\pm$ std over $3$ training seeds.}
\label{tab:lewm:latent_actions:iteration0_kl}
\centering
\footnotesize
\setlength{\tabcolsep}{3pt}
\begin{tabular}{@{}lcccc@{}}
\toprule
 & Cube & Push-T & Reacher & Two-Rooms \\
\midrule
AE & $12.55 \pm 0.43$ & $7.46 \pm 0.15$ & $3.51 \pm 0.69$ & $6.01 \pm 0.71$ \\
VAE ($\times 10^{-4}$) & $8.3 \pm 0.5$ & $3.6 \pm 3.7$ & $1.6 \pm 0.2$ & $3.9 \pm 1.4$ \\
\bottomrule
\end{tabular}
\end{table}

\emph{Without the KL term, skills do not help.}
We replace the VAE by a deterministic autoencoder (AE skills), which shares every training setting with it except the KL term. The AE model keeps the extra encoder and decoder, so if these were the source of the gains, AE skills should help as well. They do not: planning over AE skills matches raw actions on Cube and is clearly worse on the other three tasks (Figure~\ref{fig:lewm_latent_actions/success_rate_vs_distance}). What the AE lacks is alignment with the planner: a Gaussian fitted to its skills is $3.5$ to $12.6$ nats away from $\pi_0$ per skill, against less than $10^{-3}$ for VAE skills (Table~\ref{tab:lewm:latent_actions:iteration0_kl}).

\emph{Matching the mean and variance is not enough.}
The mismatch measured in Table~\ref{tab:lewm:latent_actions:iteration0_kl} is one of mean and variance, and it can be removed after training: we start CEM from the Gaussian fitted to the AE skills (``whitened'' in Figure~\ref{fig:lewm_latent_actions/success_rate_vs_distance}). If the KL term helped only by fixing these two moments, whitened AE skills would match VAE skills. They stay well below them: at the level of raw actions on Cube and Two-Rooms, and under it on Push-T and Reacher. Raw actions show the same thing: they are standardized, so they also match $\pi_0$ in mean and variance, yet VAE skills outperform them. What VAE skills add is alignment at every state. Skills can look Gaussian when pooled over the dataset while each state uses only a small part of that Gaussian. At such a state, most of the planner's samples fall where the predictor was never trained. Whitening cannot repair this, because it corrects the planner's Gaussian once, for all states. The KL term acts on each training pair. It therefore reduces the mismatch state by state, and this per-state mismatch is the one that Proposition~1 penalizes (Corollary~1).

\emph{Skills make the model accurate where the planner samples.}
We measure along whole rollouts the two prediction errors that Proposition~1 relates (Eq.~\ref{eq:prop1}): on the action chunks of the dataset, and on the planner's initial candidates (Figure~\ref{fig:pred-error-dataset-vs-planner}). If skills simply made the world model more accurate, the VAE model would have the lowest error on both. It does not: on the dataset, the three models are comparable, and the VAE model is at times the least accurate. On the planner's candidates, the errors of the raw-action and AE models are several times larger than on the dataset, while that of the VAE model grows less and is the lowest, except on Push-T beyond $40$ steps. Skills thus do not make the world model more accurate on its training data; they make it more accurate where the planner queries it.

\begin{figure}[t]
\centering
\includegraphics[width=\linewidth]{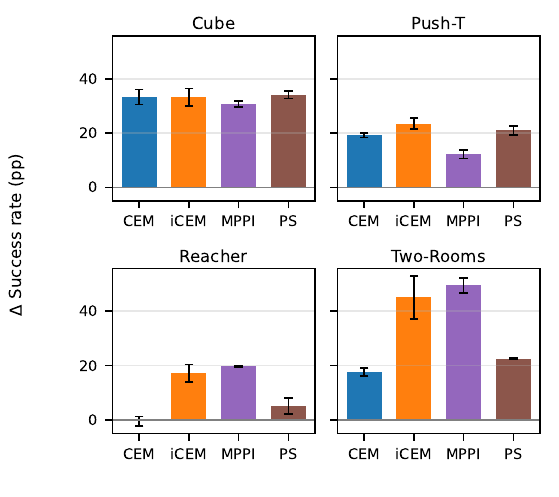}
\caption{\textbf{The gains hold for every planner.} Difference in success rate between VAE skills and raw actions, in percentage points (pp), for LeWM with four planners, averaged over the four goal distances. PS: Predictive Sampling.}
\label{fig:lewm_planner_ablation/planner_delta_success_rate_vs_distance_bars}
\end{figure}

\subsection{The gains hold across planners}

If the benefit comes from where the planner starts, it should not depend on how the planner refines its distribution afterwards. We repeat the comparison for LeWM with iCEM, MPPI and Predictive Sampling (PS). These planners start from the same $\pi_0$ as CEM, but differ from it in how they weight candidates, whether they adapt the variance, and how many iterations they run. Skills help all four: averaged over tasks and distances, the gain is $17.5$ points with CEM, $29.8$ with iCEM, $28.1$ with MPPI and $20.8$ with PS (Figure~\ref{fig:lewm_planner_ablation/planner_delta_success_rate_vs_distance_bars}). The one exception among the $16$ pairs of planner and task is CEM on Reacher, where raw actions already do well; with the other planners, skills help on Reacher too.

PS takes this test to its limit: it keeps the best of the candidates sampled from $\pi_0$ and never refines, so its gain can only come from these candidates. On Two-Rooms at $80$ steps, it reaches $94\%$ with skills, against $44\%$ with raw actions. This single round of sampling over skills even outperforms iterative CEM over raw actions in $13$ of the $16$ combinations of task and distance (Appendix~\ref{app:results}).
Changing what the planner samples thus helps more than refining how it searches. On average, skills add $21$ points to PS, while CEM's refinement adds $3$ points to raw actions.

\section{Conclusion}
\label{sec:conclusions}
\label{sec:discussions}

Planning with JEPA world models fails on distant goals partly because a sampling-based planner draws its candidates from a Gaussian, while the model has only seen the structured, state-dependent actions of its data. Skill-JEPA aligns the two with a single module: a state-conditioned VAE whose skills follow the Gaussian that the planner starts from. The planner is left untouched. Across three world models, five tasks and four planners, planning over skills reached farther goals than planning over raw actions with the same data, architectures and planner. Average success rose from $42\%$ to $62\%$ at $80$ steps on four standard benchmarks, and from $0.4\%$ to $38\%$ at two cells on AntMaze.

Controlled experiments tie the gains to this alignment. Without the KL term, skills did not help, even when the planner was adjusted to their mean and variance. With it, skills made the world model more accurate where the planner samples, not on its training data. A learned action space thus helps planning when its prior is the distribution that the planner samples from, so that the world model is trained where the planner queries it.

Our study has three limitations. First, the analysis covers only the planner's first candidates and a single prediction step. It says nothing about refinement, which did not always help: on Two-Rooms, one round of sampling over skills outperformed iterative CEM over the same skills ($94\%$ against $69\%$ at $80$ steps). Second, prediction error does not fully explain the gains. On Push-T, success improved although the model was not more accurate at long horizons (Figure~\ref{fig:pred-error-dataset-vs-planner}). Skills may also help by focusing the search on behaviors found in the data. Third, our experiments were limited to simulation, a fixed skill length, raw-action baselines and planners that start from a Gaussian.

Two directions follow. The first is to keep the search close to the skill prior during refinement, extending the alignment beyond the first iteration. The second is to combine skills with other planners, such as gradient-based ones~\cite{agrawal2025closing}, and with methods that shorten the planning problem~\citep{zhang2026hierarchicalplanninglatentworld, zhang2026hjepaendtoendlearninghierarchical}.

\section*{LLM Usage Statement}
We used Claude (Anthropic), a general-purpose LLM assistant, in the writing of this paper.
\textbf{Text.} Claude was used extensively to draft and revise text in all sections of the paper and in the appendix. Working from the authors' drafts, notes and results, it proposed wording, reorganizations and shorter versions of paragraphs, which the authors reviewed and edited.
\textbf{Framing.} We discussed with Claude how to present the work, including its positioning, scope and limitations. It suggested clarifications, remarks and checks, which the authors evaluated before adopting any of them.
\textbf{Related work.} Claude was used to search for related and concurrent work and to check our descriptions of the papers we cite. The authors read these papers and verified every reference.
\textbf{Theory.} Claude produced an initial sketch of the results in the appendix. The authors then wrote and verified the proofs, and Claude was used to check them and edit their presentation.
\textbf{Checks.} Claude was used to check the consistency of the text with the tables and figures, and the formatting and length of the manuscript.
\textbf{Code and experiments.} Claude Code, Anthropic's agentic coding tool,
was used extensively for programming throughout the project: it wrote and
modified much of the training, evaluation, result-collection and plotting code,
ran and debugged scripts, and read the code and logged run configurations to
report the hyperparameters and implementation details given in the paper. The
authors also wrote parts of the code by hand, frequently edited the generated
code, and read the code files in full. The authors specified every experiment,
reviewed the code changes, and checked results and reported values against the
code and logs. The idea, the method and the experimental design are the authors' own,
and the authors take full responsibility for the content of the paper.

{
    \small
    \bibliographystyle{ieeenat_fullname}
    \bibliography{main}

\begin{thebibliography}{46}
\providecommand{\natexlab}[1]{#1}
\providecommand{\url}[1]{\texttt{#1}}
\expandafter\ifx\csname urlstyle\endcsname\relax
  \providecommand{\doi}[1]{doi: #1}\else
  \providecommand{\doi}{doi: \begingroup \urlstyle{rm}\Url}\fi

\bibitem[Ajay et~al.(2021)Ajay, Kumar, Agrawal, Levine, and
  Nachum]{ajay2021opal}
Anurag Ajay, Aviral Kumar, Pulkit Agrawal, Sergey Levine, and Ofir Nachum.
\newblock {OPAL}: Offline primitive discovery for accelerating offline
  reinforcement learning.
\newblock In \emph{International Conference on Learning Representations}, 2021.

\bibitem[Argenson and Dulac-Arnold(2021)]{argenson2021mbop}
Arthur Argenson and Gabriel Dulac-Arnold.
\newblock Model-based offline planning.
\newblock In \emph{International Conference on Learning Representations}, 2021.

\bibitem[Assran et~al.(2025)Assran, Bardes, Fan, Garrido, Howes, Mojtaba,
  Komeili, Muckley, Rizvi, Roberts, Sinha, Zholus, Arnaud, Gejji, Martin,
  Hogan, Dugas, Bojanowski, Khalidov, Labatut, Massa, Szafraniec, Krishnakumar,
  Li, Ma, Chandar, Meier, LeCun, Rabbat, and
  Ballas]{assran2025vjepa2selfsupervisedvideo}
Mido Assran, Adrien Bardes, David Fan, Quentin Garrido, Russell Howes, Mojtaba,
  Komeili, Matthew Muckley, Ammar Rizvi, Claire Roberts, Koustuv Sinha, Artem
  Zholus, Sergio Arnaud, Abha Gejji, Ada Martin, Francois~Robert Hogan, Daniel
  Dugas, Piotr Bojanowski, Vasil Khalidov, Patrick Labatut, Francisco Massa,
  Marc Szafraniec, Kapil Krishnakumar, Yong Li, Xiaodong Ma, Sarath Chandar,
  Franziska Meier, Yann LeCun, Michael Rabbat, and Nicolas Ballas.
\newblock V-jepa 2: Self-supervised video models enable understanding,
  prediction and planning, 2025.

\bibitem[Balestriero and
  LeCun(2025)]{balestriero2025lejepaprovablescalableselfsupervised}
Randall Balestriero and Yann LeCun.
\newblock Lejepa: Provable and scalable self-supervised learning without the
  heuristics, 2025.

\bibitem[Chahe and Zhou(2026)]{chahe2026pijepa}
Amirhosein Chahe and Lifeng Zhou.
\newblock Policy-guided world model planning for language-conditioned visual
  navigation.
\newblock \emph{arXiv preprint arXiv:2603.25981}, 2026.

\bibitem[Chi et~al.(2024)Chi, Xu, Feng, Cousineau, Du, Burchfiel, Tedrake, and
  Song]{chi2024diffusionpolicyvisuomotorpolicy}
Cheng Chi, Zhenjia Xu, Siyuan Feng, Eric Cousineau, Yilun Du, Benjamin
  Burchfiel, Russ Tedrake, and Shuran Song.
\newblock Diffusion policy: Visuomotor policy learning via action diffusion,
  2024.

\bibitem[Cover and Thomas(2006)]{cover2006}
Thomas~M. Cover and Joy~A. Thomas.
\newblock \emph{Elements of Information Theory}.
\newblock Wiley-Interscience, Hoboken, NJ, 2 edition, 2006.

\bibitem[Csisz{\'a}r and K{\"o}rner(2011)]{csiszar2011}
Imre Csisz{\'a}r and J{\'a}nos K{\"o}rner.
\newblock \emph{Information Theory: Coding Theorems for Discrete Memoryless
  Systems}.
\newblock Cambridge University Press, Cambridge, UK, 2 edition, 2011.

\bibitem[Du et~al.(2026)Du, Zhang, Wang, and Wang]{du2026vlwm}
Tianqi Du, Qi Zhang, Yifei Wang, and Yisen Wang.
\newblock Beyond the next step: Variable-length latent world models for
  long-horizon planning.
\newblock \emph{arXiv preprint arXiv:2606.21775}, 2026.

\bibitem[Ha and Schmidhuber(2018)]{ha2018worldmodels}
David Ha and Jürgen Schmidhuber.
\newblock World models, 2018.

\bibitem[Hafner et~al.(2020)Hafner, Lillicrap, Ba, and
  Norouzi]{hafner2020dream}
Danijar Hafner, Timothy Lillicrap, Jimmy Ba, and Mohammad Norouzi.
\newblock Dream to control: Learning behaviors by latent imagination.
\newblock In \emph{International Conference on Learning Representations}, 2020.

\bibitem[Hansen et~al.(2024)Hansen, Su, and Wang]{hansen2024tdmpc2}
Nicklas Hansen, Hao Su, and Xiaolong Wang.
\newblock {TD-MPC2}: Scalable, robust world models for continuous control.
\newblock In \emph{International Conference on Learning Representations}, 2024.

\bibitem[Hegde et~al.(2026)Hegde, Popov, Aljalbout, and
  Smolyanskiy]{hegde2026lewam}
Shashank Hegde, Alexander Popov, Elie Aljalbout, and Nikolai Smolyanskiy.
\newblock {LeWAM}: A {JEPA} world action model with diffusion-steering-based
  {MPC}.
\newblock \emph{arXiv preprint arXiv:2610.12407}, 2026.

\bibitem[Higgins et~al.(2017)Higgins, Matthey, Pal, Burgess, Glorot, Botvinick,
  Mohamed, and Lerchner]{higgins2017betavae}
Irina Higgins, Loic Matthey, Arka Pal, Christopher Burgess, Xavier Glorot,
  Matthew Botvinick, Shakir Mohamed, and Alexander Lerchner.
\newblock beta-{VAE}: Learning basic visual concepts with a constrained
  variational framework.
\newblock In \emph{International Conference on Learning Representations}, 2017.

\bibitem[Hoffman and Johnson(2016)]{hoffman2016elbo}
Matthew~D. Hoffman and Matthew~J. Johnson.
\newblock {ELBO} surgery: yet another way to carve up the variational evidence
  lower bound.
\newblock In \emph{NIPS Workshop on Advances in Approximate Bayesian
  Inference}, 2016.

\bibitem[Howell et~al.(2022)Howell, Gileadi, Tunyasuvunakool, Zakka, Erez, and
  Tassa]{howell2022predictivesamplingrealtimebehaviour}
Taylor Howell, Nimrod Gileadi, Saran Tunyasuvunakool, Kevin Zakka, Tom Erez,
  and Yuval Tassa.
\newblock Predictive sampling: Real-time behaviour synthesis with mujoco, 2022.

\bibitem[Ivashkov et~al.(2026)Ivashkov, Balestriero, and
  Schölkopf]{ivashkov2026sensorimotorworldmodelsperception}
Petr Ivashkov, Randall Balestriero, and Bernhard Schölkopf.
\newblock Sensorimotor world models: Perception for action via inverse
  dynamics, 2026.

\bibitem[Janner et~al.(2022)Janner, Du, Tenenbaum, and
  Levine]{janner2022diffuser}
Michael Janner, Yilun Du, Joshua~B Tenenbaum, and Sergey Levine.
\newblock Planning with diffusion for flexible behavior synthesis.
\newblock \emph{arXiv preprint arXiv:2205.09991}, 2022.

\bibitem[Jiang et~al.(2023)Jiang, Zhang, Janner, Li, Rockt{\"a}schel,
  Grefenstette, and Tian]{jiang2023tap}
Zhengyao Jiang, Tianjun Zhang, Michael Janner, Yueying Li, Tim Rockt{\"a}schel,
  Edward Grefenstette, and Yuandong Tian.
\newblock Efficient planning in a compact latent action space.
\newblock In \emph{International Conference on Learning Representations}, 2023.

\bibitem[Kingma and Welling(2014)]{kingma2022autoencodingvariationalbayes}
Diederik~P Kingma and Max Welling.
\newblock Auto-encoding variational bayes, 2014.

\bibitem[LeCun(2022)]{lecun-autonomous-machine-intelligence-2022}
Yann LeCun.
\newblock A path towards autonomous machine intelligence, 2022.

\bibitem[Maes et~al.(2026{\natexlab{a}})Maes, Le~Lidec, Scieur, LeCun, and
  Balestriero]{maes_lelidec2026lewm}
Lucas Maes, Quentin Le~Lidec, Damien Scieur, Yann LeCun, and Randall
  Balestriero.
\newblock Leworldmodel: Stable end-to-end joint-embedding predictive
  architecture from pixels.
\newblock \emph{arXiv preprint}, 2026{\natexlab{a}}.

\bibitem[Maes et~al.(2026{\natexlab{b}})Maes, Lidec, Facury, Massaudi,
  Chaurasia, Capuano, Gao, Gillin, Haramati, Scieur, LeCun, and
  Balestriero]{maes2026stableworldmodelplatformreproducibleworld}
Lucas Maes, Quentin~Le Lidec, Luiz Facury, Nassim Massaudi, Ayush Chaurasia,
  Francesco Capuano, Richard Gao, Taj Gillin, Dan Haramati, Damien Scieur, Yann
  LeCun, and Randall Balestriero.
\newblock stable-worldmodel: A platform for reproducible world modeling
  research and evaluation, 2026{\natexlab{b}}.

\bibitem[Masip et~al.(2026)Masip, Swinnen, Hu, Detry, and
  Tuytelaars]{masip2026ffjepa}
Sergi Masip, Jonathan Swinnen, Yutong Hu, Renaud Detry, and Tinne Tuytelaars.
\newblock {FF-JEPA}: Long-horizon planning in world models with latent
  planners.
\newblock \emph{arXiv preprint arXiv:2606.09311}, 2026.

\bibitem[Mishra et~al.(2017)Mishra, Abbeel, and Mordatch]{mishra2017prediction}
Nikhil Mishra, Pieter Abbeel, and Igor Mordatch.
\newblock Prediction and control with temporal segment models.
\newblock In \emph{International Conference on Machine Learning}, pages
  2459--2468, 2017.

\bibitem[Murphy(2023)]{pml2Book}
Kevin~P. Murphy.
\newblock \emph{Probabilistic Machine Learning: Advanced Topics}.
\newblock MIT Press, 2023.

\bibitem[Park et~al.(2025)Park, Frans, Eysenbach, and
  Levine]{park2025ogbenchbenchmarkingofflinegoalconditioned}
Seohong Park, Kevin Frans, Benjamin Eysenbach, and Sergey Levine.
\newblock Ogbench: Benchmarking offline goal-conditioned rl, 2025.

\bibitem[Parthasarathy et~al.(2025)Parthasarathy, Kalra, Agrawal, LeCun,
  Bounou, Izmailov, and Goldblum]{agrawal2025closing}
Arjun Parthasarathy, Nimit Kalra, Rohun Agrawal, Yann LeCun, Oumayma Bounou,
  Pavel Izmailov, and Micah Goldblum.
\newblock Closing the train-test gap in world models for gradient-based
  planning.
\newblock \emph{arXiv preprint arXiv:2512.09929}, 2025.

\bibitem[Pertsch et~al.(2020)Pertsch, Lee, and
  Lim]{pertsch2020acceleratingreinforcementlearninglearned}
Karl Pertsch, Youngwoon Lee, and Joseph~J. Lim.
\newblock Accelerating reinforcement learning with learned skill priors, 2020.

\bibitem[Pinneri et~al.(2020)Pinneri, Sawant, Blaes, Achterhold, Stueckler,
  Rolinek, and Martius]{pinneri2020sampleefficientcrossentropymethodrealtime}
Cristina Pinneri, Shambhuraj Sawant, Sebastian Blaes, Jan Achterhold, Joerg
  Stueckler, Michal Rolinek, and Georg Martius.
\newblock Sample-efficient cross-entropy method for real-time planning, 2020.

\bibitem[Rubinstein(1999)]{Rubinstein1999}
Reuven Rubinstein.
\newblock The cross-entropy method for combinatorial and continuous
  optimization.
\newblock \emph{Methodology and Computing in Applied Probability}, 1\penalty0
  (2):\penalty0 127--190, 1999.

\bibitem[Shi et~al.(2022)Shi, Lim, and Lee]{shi2022skimo}
Lucy~Xiaoyang Shi, Joseph~J Lim, and Youngwoon Lee.
\newblock Skill-based model-based reinforcement learning.
\newblock \emph{arXiv preprint arXiv:2207.07560}, 2022.

\bibitem[Sobal et~al.(2025)Sobal, Zhang, Cho, Balestriero, Rudner, and
  LeCun]{sobal2025learningrewardfreeofflinedata}
Vlad Sobal, Wancong Zhang, Kyunghyun Cho, Randall Balestriero, Tim G.~J.
  Rudner, and Yann LeCun.
\newblock Learning from reward-free offline data: A case for planning with
  latent dynamics models, 2025.

\bibitem[Sohn et~al.(2015)Sohn, Lee, and Yan]{sohn2015cvae}
Kihyuk Sohn, Honglak Lee, and Xinchen Yan.
\newblock Learning structured output representation using deep conditional
  generative models.
\newblock In \emph{Advances in Neural Information Processing Systems}, 2015.

\bibitem[Sutton and Barto(2018)]{Sutton1998}
Richard~S. Sutton and Andrew~G. Barto.
\newblock \emph{Reinforcement Learning: An Introduction}.
\newblock The MIT Press, second edition, 2018.

\bibitem[Tassa et~al.(2018)Tassa, Doron, Muldal, Erez, Li, de~Las~Casas,
  Budden, Abdolmaleki, Merel, Lefrancq, Lillicrap, and
  Riedmiller]{tassa2018deepmindcontrolsuite}
Yuval Tassa, Yotam Doron, Alistair Muldal, Tom Erez, Yazhe Li, Diego de
  Las~Casas, David Budden, Abbas Abdolmaleki, Josh Merel, Andrew Lefrancq,
  Timothy Lillicrap, and Martin Riedmiller.
\newblock Deepmind control suite, 2018.

\bibitem[Terver et~al.(2025)Terver, Yang, Ponce, Bardes, and
  LeCun]{terver2025jepawms}
Basile Terver, Tsung-Yen Yang, Jean Ponce, Adrien Bardes, and Yann LeCun.
\newblock What drives success in physical planning with joint-embedding
  predictive world models?
\newblock \emph{arXiv preprint arXiv:2512.24497}, 2025.

\bibitem[Wagener et~al.(2019)Wagener, Cheng, Sacks, and
  Boots]{wagener2019onlinelearningapproachmodel}
Nolan Wagener, Ching-An Cheng, Jacob Sacks, and Byron Boots.
\newblock An online learning approach to model predictive control, 2019.

\bibitem[Williams et~al.(2017)Williams, Drews, Goldfain, Rehg, and
  Theodorou]{williams2017informationtheoreticmodelpredictive}
Grady Williams, Paul Drews, Brian Goldfain, James~M. Rehg, and Evangelos~A.
  Theodorou.
\newblock Information theoretic model predictive control: Theory and
  applications to autonomous driving, 2017.

\bibitem[You et~al.(2026)You, Zhang, Ran, Yang, Zhang, Xue, Song, Tian, and
  Guo]{you2026control}
Hanzhe You, Yonggang Zhang, Maohao Ran, Zhiqin Yang, Zhenyuan Zhang, Wei Xue,
  Jun Song, Xinmei Tian, and Yike Guo.
\newblock A control theory of predictability in latent world models.
\newblock \emph{arXiv preprint arXiv:2607.10362}, 2026.

\bibitem[Yu et~al.(2020)Yu, Thomas, Yu, Ermon, Zou, Levine, Finn, and
  Ma]{yu2020mopo}
Tianhe Yu, Garrett Thomas, Lantao Yu, Stefano Ermon, James~Y Zou, Sergey
  Levine, Chelsea Finn, and Tengyu Ma.
\newblock {MOPO}: Model-based offline policy optimization.
\newblock \emph{Advances in neural information processing systems},
  33:\penalty0 14129--14142, 2020.

\bibitem[Zhang et~al.(2026{\natexlab{a}})Zhang, Terver, Rabbat, LeCun, and
  Balestriero]{zhang2026hjepaendtoendlearninghierarchical}
Wancong Zhang, Basile Terver, Michael Rabbat, Yann LeCun, and Randall
  Balestriero.
\newblock H-jepa: End-to-end learning of hierarchical world models for visual
  planning, 2026{\natexlab{a}}.

\bibitem[Zhang et~al.(2026{\natexlab{b}})Zhang, Terver, Zholus, Chitnis,
  Sutaria, Assran, Balestriero, Bar, LeCun, and
  Ballas]{zhang2026hierarchicalplanninglatentworld}
Wancong Zhang, Basile Terver, Artem Zholus, Soham Chitnis, Harsh Sutaria, Mido
  Assran, Randall Balestriero, Amir Bar, Yann LeCun, and Nicolas Ballas.
\newblock Hierarchical planning with latent world models, 2026{\natexlab{b}}.

\bibitem[Zhao et~al.(2023)Zhao, Kumar, Levine, and
  Finn]{zhao2023learningfinegrainedbimanualmanipulation}
Tony~Z. Zhao, Vikash Kumar, Sergey Levine, and Chelsea Finn.
\newblock Learning fine-grained bimanual manipulation with low-cost hardware,
  2023.

\bibitem[Zhou et~al.(2025)Zhou, Pan, LeCun, and
  Pinto]{zhou2025dinowmworldmodelspretrained}
Gaoyue Zhou, Hengkai Pan, Yann LeCun, and Lerrel Pinto.
\newblock Dino-wm: World models on pre-trained visual features enable zero-shot
  planning, 2025.

\bibitem[Zhou et~al.(2021)Zhou, Bajracharya, and Held]{zhou2020plas}
Wenxuan Zhou, Sujay Bajracharya, and David Held.
\newblock {PLAS}: Latent action space for offline reinforcement learning.
\newblock In \emph{Conference on Robot Learning}, pages 1719--1735, 2021.

\end{thebibliography}
}

\clearpage
\setcounter{page}{1}
\maketitlesupplementary

\appendix

\newcommand{\proofhead}{\par\smallskip\noindent\textbf{Proof.}\ }
\newcommand{\proofstep}[2]{\par\medskip\noindent\textbf{#1} (#2).\ }
\newcommand{\qedbox}{\hfill$\square$\par\medskip}

\section{Proofs}
\label{app:proofs}

\paragraph{Setting.}
We recall the objects from Sections~\ref{sec:skill-vae} and~\ref{sec:prior-mismatch}.
A state $s$ is encoded into a latent observation $z = f_\theta(s)$. An action chunk $a$
is encoded into a skill $l \in \mathbb{R}^d$ by the conditional encoder
$q_\theta(l \mid a, z)$, trained against a fixed prior $p(l) = \mathcal{N}(0, I)$. The
planner draws its initial candidates from a distribution $\pi_0$ over skills. For a
state $s$, the per-state aggregate posterior is the mixture of encoder outputs over the
actions taken from $s$ in the data,
\begin{equation*}
  \bar q_z(l) \;=\; q_{\text{agg}}(l \mid z) \;=\; \mathbb{E}_{a \sim p_{\text{data}}(a \mid s)}\!\left[q_\theta(l \mid a, z)\right],
\end{equation*}
which is a density since it is a mixture of densities. It is the distribution of skills
seen at $s$ during training. Let $e(s, l)$ be the error the predictor makes when skill $l$ is used at state $s$:

\begin{equation*}
e(s, l) = \mathbb{E}_{s'}\bigl\|\, g_\theta(z, l) - f_\theta(s') \,\bigr\|_2
\end{equation*}

where $z = f_\theta(s)$ and $s'$ is the state reached from $s$ by executing the decoded chunk $\hat a_\theta(l, z)$ for $k$ steps. The error under $\bar q_z$ therefore includes the decoder's reconstruction error as well as the predictor's, which is the relevant quantity for planning.
The error is bounded, $0 \le e \le B$ for some $B > 0$. This holds for the trained model at planning time: both
$f_\theta$ and $g_\theta$ end in a LayerNorm followed by fixed affine maps and
GELU (BatchNorm uses running statistics in evaluation mode), so
$\|z\|, \|g_\theta(z, l)\| \le R$ for a constant $R$ determined by the trained
weights, uniformly in $l$, and hence $B = 2R$.

\paragraph{Results.}
Proposition~1 bounds the predictor's expected error under the planner's initial
distribution $\pi_0$ by its error under $\bar q_z$ plus a mismatch term,
$B \sqrt{\tfrac{1}{2}\mathrm{KL}(\bar q_z \,\|\, \pi_0)}$. Corollary~1 shows that, when
$\pi_0 = p$, the mismatch term averaged over states is controlled by the KL term of the
training objective. Both rely on two standard facts, Lemmas~1 and~2.

\paragraph{Notation.}
$P$ and $Q$ denote probability measures on $\mathbb{R}^d$ with densities $\rho_P$ and
$\rho_Q$. The total variation distance is
$\mathrm{TV}(P, Q) = \sup_{S} |P(S) - Q(S)|$, with the supremum over measurable sets,
and $\mathrm{KL}(P \,\|\, Q) = \int \rho_P \log (\rho_P / \rho_Q)\,dx$. Logarithms are
natural, so divergences are in nats.

\paragraph{Lemma 1 (bounded functions under total variation).}
Let $f: \mathbb{R}^d \to [0, B]$ be measurable. Then
\begin{equation*}
  \bigl|\mathbb{E}_P[f] - \mathbb{E}_Q[f]\bigr| \;\le\; B\, \mathrm{TV}(P, Q).
\end{equation*}

\proofhead Let $A = \{x : \rho_P(x) > \rho_Q(x)\}$. Since both densities integrate
to one, $\int (\rho_P - \rho_Q)\,dx = 0$, and therefore
\begin{align*}
  \int_A (\rho_P - \rho_Q)\,dx &= \int_{A^c} (\rho_Q - \rho_P)\,dx \\
                               &= P(A) - Q(A) \;\le\; \mathrm{TV}(P, Q).
\end{align*}
Splitting the domain into $A$ and $A^c$,
\begin{align*}
  \mathbb{E}_P[f] - \mathbb{E}_Q[f]
  &= \int_A f\,(\rho_P - \rho_Q)\,dx \\
  &\quad - \int_{A^c} f\,(\rho_Q - \rho_P)\,dx \\[4pt]
  &\le \int_A f\,(\rho_P - \rho_Q)\,dx \\[4pt]
  &\le B \int_A (\rho_P - \rho_Q)\,dx \\[4pt]
  &\le B\, \mathrm{TV}(P, Q).
\end{align*}
The first inequality drops the subtracted term, which is non-negative, since $f \ge 0$ and $\rho_Q \ge \rho_P$ on
$A^c$. The second uses $f \le B$ and $\rho_P - \rho_Q > 0$ on $A$. The last uses the
identity above. Exchanging the roles of $P$ and $Q$ gives the same bound for
$\mathbb{E}_Q[f] - \mathbb{E}_P[f]$, which proves the claim. \qedbox

\paragraph{Lemma 2 (Pinsker's inequality).}
For any probability measures $P$ and $Q$,
\begin{equation*}
  \mathrm{TV}(P, Q) \;\le\; \sqrt{\tfrac{1}{2}\,\mathrm{KL}(P \,\|\, Q)}
\end{equation*}
\citep[Lemma~11.6.1]{cover2006}.\footnote{Cover \& Thomas state the discrete case, with the $L_1$ norm and base-2 logarithms; since $\|P - Q\|_1 = 2\,\mathrm{TV}(P, Q)$, their statement in nats is the one above;
for general measurable spaces see \citet{csiszar2011}.}
Since TV is symmetric, either direction of the KL may be used.

\paragraph{Proposition 1.}
For every state $s$, with $z = f_\theta(s)$,
\begin{equation*}
  \mathbb{E}_{\pi_0}[e(s, l)]
  \;\le\; \mathbb{E}_{\bar q_z}[e(s, l)]
  + B \sqrt{\tfrac{1}{2}\,\mathrm{KL}(\bar q_z \,\|\, \pi_0)}.
\end{equation*}

\proofhead Fix $s$. Lemma~1 with $f = e(s, \cdot)$, $P = \pi_0$ and $Q = \bar q_z$
gives
\begin{equation*}
  \mathbb{E}_{\pi_0}[e(s, l)]
  \;\le\; \mathbb{E}_{\bar q_z}[e(s, l)] + B\, \mathrm{TV}(\bar q_z, \pi_0).
\end{equation*}
Lemma~2 bounds the total variation term:
\begin{equation*}
  \mathrm{TV}(\bar q_z, \pi_0)
  \;\le\; \sqrt{\tfrac{1}{2}\,\mathrm{KL}(\bar q_z \,\|\, \pi_0)}.
\end{equation*}
Substituting the second inequality into the first proves the claim. \qedbox

\paragraph{Corollary 1 (control by the training objective).}
Let $\pi_0 = p$, and denote the KL term of the training objective by
\begin{equation*}
  \mathcal{K} \;=\; \mathbb{E}_{(s, a) \sim p_{\text{data}}}\!\left[\mathrm{KL}\!\left(q_\theta(l \mid a, z)\,\|\,p\right)\right],
  \qquad z = f_\theta(s).
\end{equation*}
Then, averaging over states $s \sim p_{\text{data}}$,
\begin{equation*}
  \mathbb{E}_{s}\,\mathbb{E}_{\pi_0}[e(s, l)]
  \;\le\; \mathbb{E}_{s}\,\mathbb{E}_{\bar q_z}[e(s, l)] + B \sqrt{\tfrac{1}{2}\,\mathcal{K}}.
\end{equation*}

\proofhead Let $\mathrm{KL}_s = \mathrm{KL}(\bar q_z \,\|\, p)$.

\proofstep{Step 1}{averaging} Averaging Proposition~1 over $s$ gives
\begin{equation*}
  \mathbb{E}_{s}\,\mathbb{E}_{\pi_0}[e(s, l)]
  \;\le\; \mathbb{E}_{s}\,\mathbb{E}_{\bar q_z}[e(s, l)]
  + B\, \mathbb{E}_{s}\!\left[\sqrt{\tfrac{1}{2}\,\mathrm{KL}_s}\right].
\end{equation*}
Since the square root is concave, Jensen's inequality gives
\begin{equation*}
  \mathbb{E}_{s}\!\left[\sqrt{\tfrac{1}{2}\,\mathrm{KL}_s}\right]
  \;\le\; \sqrt{\tfrac{1}{2}\,\mathbb{E}_{s}[\mathrm{KL}_s]}.
\end{equation*}
It therefore suffices to show $\mathbb{E}_{s}[\mathrm{KL}_s] \le \mathcal{K}$.

\proofstep{Step 2}{decomposition at a fixed state} Fix $s$ and write
$q_a = q_\theta(\cdot \mid a, z)$, so that $\mathbb{E}_{a \mid s}[q_a] = \bar q_z$ by
definition. Multiplying and dividing by $\bar q_z$ inside the logarithm,
\begin{align*}
  \mathbb{E}_{a \mid s}\!\left[\mathrm{KL}(q_a \,\|\, p)\right]
  &= \mathbb{E}_{a \mid s}\!\int q_a(l) \log \frac{q_a(l)}{\bar q_z(l)}\, dl \\[4pt]
  &\quad + \mathbb{E}_{a \mid s}\!\int q_a(l) \log \frac{\bar q_z(l)}{p(l)}\, dl.
\end{align*}
The first term is $\mathbb{E}_{a \mid s}[\mathrm{KL}(q_a \,\|\, \bar q_z)] = I_q(l; a \mid s) \ge 0$,
the conditional mutual information at state $s$. In the second term the logarithm does
not depend on $a$, so exchanging expectation and integral,
\begin{align*}
  \int \mathbb{E}_{a \mid s}[q_a(l)] \log \frac{\bar q_z(l)}{p(l)}\, dl
  &= \int \bar q_z(l) \log \frac{\bar q_z(l)}{p(l)}\, dl \\[4pt]
  &= \mathrm{KL}_s.
\end{align*}
Hence
\begin{equation*}
  \mathbb{E}_{a \mid s}\!\left[\mathrm{KL}(q_a \,\|\, p)\right]
  \;=\; I_q(l; a \mid s) + \mathrm{KL}_s \;\ge\; \mathrm{KL}_s.
\end{equation*}

\proofstep{Step 3}{conclusion} Averaging the identity of Step~2 over $s \sim p_{\text{data}}$,
the left-hand side becomes $\mathcal{K}$, since
$\mathbb{E}_{s}\,\mathbb{E}_{a \mid s} = \mathbb{E}_{(s, a) \sim p_{\text{data}}}$.

This is a per-state version of the decomposition of \citet{hoffman2016elbo}:
\begin{equation*}
  \mathcal{K} \;=\; \mathbb{E}_{s}\!\left[I_q(l; a \mid s)\right] + \mathbb{E}_{s}[\mathrm{KL}_s]
  \;\ge\; \mathbb{E}_{s}[\mathrm{KL}_s].
\end{equation*}
Combining Step~1 with this inequality, and using that the square root is increasing,
\begin{align*}
  \mathbb{E}_{s}\,\mathbb{E}_{\pi_0}[e(s, l)]
  &\le \mathbb{E}_{s}\,\mathbb{E}_{\bar q_z}[e(s, l)]
     + B \sqrt{\tfrac{1}{2}\,\mathbb{E}_{s}[\mathrm{KL}_s]} \\[4pt]
  &\le \mathbb{E}_{s}\,\mathbb{E}_{\bar q_z}[e(s, l)]
     + B \sqrt{\tfrac{1}{2}\,\mathcal{K}},
\end{align*}
which is the claim. \qedbox

\paragraph{Remark (the prior need not be Gaussian).}
Neither result uses the form of the prior. Proposition~1 holds for any initial distribution $\pi_0$, and the proof of Corollary~1 goes through for any fixed prior $p$ with $\pi_0 = p$. The standard Gaussian is the choice imposed by most of the planners we use (Section~\ref{sec:planners}); it is not required by the analysis. CEM, MPPI and Predictive Sampling draw their first candidates from $\mathcal{N}(0, \sigma_0^2 I)$ with $\sigma_0 = 1$ in all our runs, so $\pi_0 = p$ and Corollary~1 applies directly. iCEM draws temporally correlated noise, normalized to unit scale along the horizon, so its per-step marginals have roughly unit scale but its joint initial distribution is not $\mathcal{N}(0, I)$. Proposition~1 still holds for iCEM, with $\pi_0$ its colored-noise distribution, but Corollary~1 does not apply to it as stated. Since every planner also includes the current mean as a candidate, one of the $K$ initial candidates is $l = 0$ and the remaining $K - 1$ are drawn from $\pi_0$. The bound also concerns where a planner starts, not how it searches, so it applies equally to a planner that refines Gaussian-initialized candidates by gradient descent. Finally, total variation and KL divergence are invariant under invertible maps. If a planner draws its first candidates as $u = T(\epsilon)$, with $\epsilon \sim \mathcal{N}(0, I)$ and $T$ a known invertible map (for uniform proposals on $[0,1]^d$, $T$ is the Gaussian CDF applied coordinate-wise), then planning over $u$ and feeding the model $l = T^{-1}(u)$ leaves every term of Proposition~1 unchanged. Such planners are covered without retraining.

\section{Skill objective}
\label{app:skill-objective}

This section derives the action terms of the training objective (Eq.~\eqref{eq:objective}).

\paragraph{Data.}
From each trajectory we extract all pairs $(s_t, a_{t:t+k-1})$ and pool them over trajectories and start times. This defines a time-independent data distribution $p_{\text{data}}(s, a)$ of action chunks, where we write $a$ for the chunk $a_{t:t+k-1}$ and drop the time index. The context of an action chunk is the latent observation $z = f_\theta(s)$. Since all modules are trained jointly without stopping gradients, $z$ depends on $\theta$, so we define the data distribution over states rather than latents.

\paragraph{Objective.}
For the variational objective we interpret the decoder as the mean of a Gaussian likelihood with a fixed standard deviation $\sigma_a$, $p_\theta(a \mid l, z) = \mathcal{N}(\hat a_\theta(l, z), \sigma_a^2 I)$. For each pair $(s, a)$, with $z = f_\theta(s)$, the conditional ELBO lower-bounds the conditional log-likelihood \citep[Ch.~21]{pml2Book}:

\begin{equation}
\begin{split}
  \log p_\theta(a \mid z) \;\geq\;
  &\mathbb{E}_{q_\theta(l \mid a, z)}\!\left[\log p_\theta(a \mid l, z)\right] \\
  &- \mathrm{KL}\!\left(q_\theta(l \mid a, z)\,\|\,p(l)\right).
\end{split}
\end{equation}

Following \citet{higgins2017betavae} we weight the KL term by $\beta$ and maximize the expected objective over the data distribution:

\begin{equation}
\begin{split}
  \mathcal{J}^{\beta}(\theta) =
  \mathbb{E}_{(s, a) \sim p_{\text{data}}}\Big[
    &\mathbb{E}_{q_\theta(l \mid a, z)}\!\left[\log p_\theta(a \mid l, z)\right] \\
    &- \beta\, \mathrm{KL}\!\left(q_\theta(l \mid a, z)\,\|\,p(l)\right)
  \Big].
\end{split}
\label{eq:beta-elbo}
\end{equation}

For $\beta = 1$ this is the expected ELBO. For $\beta < 1$ the objective is no longer a lower bound but a reconstruction objective regularized toward the prior. In practice the outer expectation is estimated with minibatches $\{(s_i, a_i)\}_{i=1}^{N}$, $z_i = f_\theta(s_i)$, and the inner one with a single reparameterized sample $l_i \sim q_\theta(l \mid a_i, z_i)$ \citep{kingma2022autoencodingvariationalbayes}. Under this likelihood, $-\log p_\theta(a \mid l, z) = \frac{1}{2\sigma_a^2}\|\hat a_\theta(l, z) - a\|^2 + \text{const}$, so maximizing $\mathcal{J}^\beta$ is equivalent to minimizing $\mathcal{L}_{\text{rec}} + 2\sigma_a^2\beta\,\mathcal{L}_{\text{KL}}$. Since $\sigma_a$ is not learned, we treat the product as a single weight and write $\beta$ for it in the full objective (Eq.~\eqref{eq:objective}).

\section{Planning algorithm}
\label{app:planning-algos}

Algorithm~\ref{alg:sampling-planner} gives the template shared by all the planners we evaluate (Section~\ref{sec:planners}).

\begin{algorithm}[ht]
\caption{Sampling-based planning over skill latents (one replanning step). Our
  default planner is CEM, with $h \equiv 1$, $\alpha = 1$ and $\sigma$ refit; the other
  planners we evaluate are instances of the same template with different settings
  (Appendix~\ref{app:eval-details}).}
\label{alg:sampling-planner}
\begin{algorithmic}[1]
\Require state $s_t$, goal $s_g$, horizon of $M$ skills ($H = Mk$ steps), initial scale $\sigma_0$, candidates $K$,
  elites $E$, iterations $I$, weighting $h$, step size $\alpha$, commitment $C$
\LineComment{Initialize at the prior; encode observations}
\State $\mu_{1:M} \gets 0$, \quad $\sigma_{1:M} \gets \sigma_0$
\State $z_t \gets f_\theta(s_t)$, \quad $z_g \gets f_\theta(s_g)$
\For{$i = 1, \dots, I$}
  \LineComment{Sample candidates, always including the mean}
  \State $l^{(1)}_{1:M} \gets \mu_{1:M}$, \quad $l^{(n)}_{1:M} \sim \mathcal{N}\big(\mu_{1:M}, \mathrm{diag}\,\sigma^2_{1:M}\big)$ for $n = 2, \dots, K$
  \LineComment{Roll out and score each candidate}
  \State $\hat z^{(n)}_{t} \gets z_t$
  \State $\hat z^{(n)}_{t+jk} \gets g_\theta\big(\hat z^{(n)}_{t+(j-1)k},\, l^{(n)}_j\big)$ for $j = 1, \dots, M$
  \State $J^{(n)} \gets \lVert \hat z^{(n)}_{t+H} - z_g \rVert_2^2$
  \LineComment{Select elites; refit $\mu$, and $\sigma$ if refit}
  \State $\mathcal{E} \gets$ indices of the $E$ lowest costs
  \State $w^{(n)} \gets h(J^{(n)}) \,/\, \sum_{m \in \mathcal{E}} h(J^{(m)})$ for $n \in \mathcal{E}$
  \State $\mu_{\mathcal{E}} \gets \sum_{n \in \mathcal{E}} w^{(n)} l^{(n)}$, \quad $\sigma^2_{\mathcal{E}} \gets \sum_{n \in \mathcal{E}} w^{(n)} (l^{(n)} - \mu_{\mathcal{E}})^2$
  \State $\mu \gets (1-\alpha)\,\mu + \alpha\,\mu_{\mathcal{E}}$, \quad $\sigma^2 \gets (1-\alpha)\,\sigma^2 + \alpha\,\sigma^2_{\mathcal{E}}$
\EndFor
\LineComment{Decode the plan along its rollout}
\State $l^\star_{1:M} \gets \mu_{1:M}$, \quad $\hat z^\star_{t} \gets z_t$
\For{$j = 1, \dots, M$}
  \State $\hat a_{t+(j-1)k:\,t+jk-1} \gets \hat a_\theta\big(l^\star_j,\, \hat z^\star_{t+(j-1)k}\big)$
  \State $\hat z^\star_{t+jk} \gets g_\theta\big(\hat z^\star_{t+(j-1)k},\, l^\star_j\big)$
\EndFor
\LineComment{Execute open loop and replan}
\State execute $\hat a_{t:t+C-1}$; replan from $s_{t+C}$
\end{algorithmic}
\end{algorithm}

\section{Evaluation details}
\label{app:eval-details}

All planners we evaluate are instances of Algorithm~\ref{alg:sampling-planner}. They differ in the candidate count $K$, the elite count $E$, the weighting $h$ applied to elite costs,
whether $\sigma$ is refit, the step size $\alpha$, the number of inner iterations
$I$, and the noise (Table~\ref{tab:planners}).
CEM \citep{Rubinstein1999} weights the $E$ lowest-cost samples uniformly and refits
both moments. iCEM \citep{pinneri2020sampleefficientcrossentropymethodrealtime} keeps
the same weighting but samples temporally correlated (colored) noise, carries elites
across iterations, and updates the moments with momentum $\alpha < 1$.
MPPI \citep{williams2017informationtheoreticmodelpredictive} uses exponential weights
with temperature $\lambda$ over all $K$ candidates and keeps $\sigma$ fixed.
Predictive Sampling (PS) \citep{howell2022predictivesamplingrealtimebehaviour} is the
limiting case $E = 1$, equivalently $\lambda \to 0$, with fixed $\sigma$ and a single
iteration. Without warm starting it reduces to one-shot random shooting: the best of
$K - 1$ samples from $\mathcal{N}(0, \sigma_0^2 I)$ and the prior mean.

All implementations are from stable-worldmodel \citep{maes2026stableworldmodelplatformreproducibleworld}. One detail differs from the original formulations: every planner includes the current mean as a candidate at each iteration.

\begin{table}[ht]
\centering
\caption{Planners as instances of Algorithm~\ref{alg:sampling-planner}, with the values used in all experiments. All planners start from $\mu = 0$, $\sigma_0 = 1$ at every replanning step. MPPI uses $\lambda = 0.5$; iCEM uses colored noise with exponent $2$.}
\label{tab:planners}
\begin{tabular}{@{}lcccc@{}}
\toprule
 & CEM & iCEM & MPPI & PS \\
\midrule
Candidates $K$     & 300        & 300        & 300                & 9000  \\
Elites $E$         & 30         & 30         & all                & 1     \\
Weighting $h(c)$   & $1$        & $1$        & $\exp(-c/\lambda)$ & $1$   \\
$\sigma$ refit     & \checkmark & \checkmark & --                 & --    \\
Step size $\alpha$ & 1          & 0.9        & 1                  & 1     \\
Iterations $I$     & 30         & 30         & 30                 & 1     \\
Noise              & white      & colored    & white              & white \\
Elite carry-over   & --         & 5          & --                 & --    \\
\bottomrule
\end{tabular}
\end{table}

\paragraph{Execution.}
Plans are executed open loop in full ($C = H$) and the planner replans once the plan is exhausted, with a budget of twice the goal offset in environment steps, so each episode uses at most two plans. The chunk length is $k = 4$ for all tasks, so the plan horizon is $M = \text{offset}/4$ skills.

\paragraph{Training.}
The skill encoder and decoder are trained jointly with the world model for 10 epochs with AdamW ($\beta_1 = 0.9$, $\beta_2 = 0.999$, weight decay $10^{-3}$, none on biases and normalization parameters), batch size 128, peak learning rate $5 \times 10^{-5}$ with linear warmup over the first 1\% of steps and cosine decay to 0, in bf16 mixed precision. The KL weight in Eq.~\eqref{eq:objective} is $\beta = 0.01$, with the KL summed over latent dimensions. The skill encoder embeds each of the $k$ actions with a linear layer, LayerNorm and GELU, prepends a CLS token and an observation token, and applies a 2-layer bidirectional Transformer (width 192, 4 heads, MLP width 768) without positional embeddings; the CLS output gives the mean and log-variance of $q_\theta$. Without positional embeddings the skill is invariant to the order of actions within a chunk; the decoder produces an ordered chunk from $k$ learned query tokens and an observation token, processed by 2 adaLN-Zero Transformer blocks conditioned on a linear projection of $l$, followed by LayerNorm and a linear output layer. Table~\ref{tab:hparams-task} lists the settings that vary across tasks.

\begin{table}[ht]
\centering
\footnotesize
\caption{Per-task settings. AntMaze goal offsets of 32, 52, 64 and 84 steps correspond to 1--4 maze cells.}
\label{tab:hparams-task}
\setlength{\tabcolsep}{4pt}
\begin{tabular}{@{}lccccc@{}}
\toprule
 & Cube & Push-T & Reacher & Two-Rooms & AntMaze \\
\midrule
Action dim          & 5      & 2      & 2      & 2      & 8      \\
Skill dim $d$       & 10     & 4      & 4      & 4      & 16     \\
Episodes            & 10k    & 18.7k  & 10k    & 10k    & 25k    \\
Training steps      & 131k   & 145k   & 131k   & 54k    & 680k   \\
Goal offsets        & 20--80 & 20--80 & 20--80 & 20--80 & 32--84 \\
Plan horizon $M$    & 5--20  & 5--20  & 5--20  & 5--20  & 8--21  \\
\bottomrule
\end{tabular}
\end{table}

\section{Results}
\label{app:results}

This section reports the full results summarized in the main paper. Unless stated otherwise, success rates are percentages over 100 evaluation episodes, reported as mean $\pm$ standard deviation over 3 training seeds. Goal distances are given in environment steps for the four standard tasks and in maze cells for AntMaze. In all tables, bold marks the highest mean among the rows being compared, for each task and goal distance.

\subsection{Skill-JEPA across world models}

Table~\ref{tab:methods:all:success_rate_vs_distance} reports success rates for raw actions and VAE skills with each of the three world models (LeWM, PLDM, SMWM) on the four standard tasks, planned with CEM. Table~\ref{tab:mean_success_vs_distance} averages these results over tasks and world models. Averaged over all pairs, the advantage of VAE skills grows from 4.9 points at 20 steps to 20.4 points at 80 steps. At 80 steps, VAE skills have a higher mean success rate in 10 of the 12 task and world model pairs. The two exceptions are LeWM and SMWM on Reacher, where raw actions are ahead by 1 and 6 points, respectively.

\begin{table*}[t]
\caption{Success rate (\%) vs.\ goal distance (environment steps) for raw actions and VAE skills across world models, planned with CEM. Mean $\pm$ std over 3 training seeds.}
\label{tab:methods:all:success_rate_vs_distance}

\centering
\footnotesize
\renewcommand{\arraystretch}{1.3}
\begin{tabular}{llcccc}
\toprule
 &  & cube (20\,/\,40\,/\,60\,/\,80) & pusht (20\,/\,40\,/\,60\,/\,80) & reacher (20\,/\,40\,/\,60\,/\,80) & tworoom (20\,/\,40\,/\,60\,/\,80) \\
\midrule
\multirow[t]{2}{*}{lewm} & raw & 80$\pm$1\,/\,50$\pm$2\,/\,41$\pm$1\,/\,47$\pm$2 & \textbf{96$\pm$2}\,/\,64$\pm$3\,/\,24$\pm$6\,/\,6$\pm$2 & \textbf{84$\pm$2}\,/\,81$\pm$4\,/\,\textbf{75$\pm$6}\,/\,\textbf{70$\pm$3} & \textbf{90$\pm$4}\,/\,72$\pm$4\,/\,55$\pm$6\,/\,32$\pm$1 \\
 & vae & \textbf{98$\pm$1}\,/\,\textbf{92$\pm$3}\,/\,\textbf{79$\pm$6}\,/\,\textbf{82$\pm$2} & 94$\pm$2\,/\,\textbf{82$\pm$4}\,/\,\textbf{59$\pm$1}\,/\,\textbf{33$\pm$4} & \textbf{84$\pm$2}\,/\,\textbf{82$\pm$3}\,/\,73$\pm$3\,/\,69$\pm$5 & 90$\pm$0\,/\,\textbf{84$\pm$1}\,/\,\textbf{77$\pm$1}\,/\,\textbf{69$\pm$3} \\
\cline{1-6}
\multirow[t]{2}{*}{pldm} & raw & 66$\pm$1\,/\,46$\pm$3\,/\,35$\pm$2\,/\,47$\pm$2 & 80$\pm$3\,/\,46$\pm$5\,/\,18$\pm$2\,/\,7$\pm$2 & \textbf{83$\pm$1}\,/\,\textbf{78$\pm$5}\,/\,\textbf{71$\pm$6}\,/\,57$\pm$7 & \textbf{97$\pm$2}\,/\,\textbf{88$\pm$3}\,/\,\textbf{81$\pm$5}\,/\,60$\pm$4 \\
 & vae & \textbf{97$\pm$1}\,/\,\textbf{72$\pm$2}\,/\,\textbf{71$\pm$2}\,/\,\textbf{67$\pm$4} & \textbf{93$\pm$1}\,/\,\textbf{75$\pm$3}\,/\,\textbf{43$\pm$2}\,/\,\textbf{20$\pm$3} & 80$\pm$3\,/\,78$\pm$3\,/\,70$\pm$7\,/\,\textbf{61$\pm$7} & 92$\pm$1\,/\,86$\pm$1\,/\,80$\pm$2\,/\,\textbf{75$\pm$2} \\
\cline{1-6}
\multirow[t]{2}{*}{smwm} & raw & 95$\pm$2\,/\,61$\pm$2\,/\,55$\pm$2\,/\,59$\pm$2 & \textbf{89$\pm$2}\,/\,62$\pm$5\,/\,28$\pm$3\,/\,9$\pm$1 & 83$\pm$3\,/\,\textbf{81$\pm$5}\,/\,\textbf{73$\pm$6}\,/\,\textbf{66$\pm$5} & 97$\pm$3\,/\,83$\pm$15\,/\,63$\pm$23\,/\,41$\pm$30 \\
 & vae & \textbf{98$\pm$1}\,/\,\textbf{96$\pm$2}\,/\,\textbf{97$\pm$2}\,/\,\textbf{98$\pm$2} & 87$\pm$4\,/\,\textbf{69$\pm$4}\,/\,\textbf{41$\pm$4}\,/\,\textbf{21$\pm$3} & \textbf{85$\pm$3}\,/\,79$\pm$1\,/\,71$\pm$5\,/\,60$\pm$6 & \textbf{100$\pm$0}\,/\,\textbf{98$\pm$4}\,/\,\textbf{97$\pm$4}\,/\,\textbf{92$\pm$11} \\
\cline{1-6}
\bottomrule
\end{tabular}
\end{table*}

\begin{table}[t]
\caption{Mean success rate (\%) vs.\ goal distance (environment steps), averaged over the four tasks and three world models of Table~\ref{tab:methods:all:success_rate_vs_distance}.}
\label{tab:mean_success_vs_distance}
\centering
\footnotesize
\renewcommand{\arraystretch}{1.3}
\begin{tabular}{lcccc}
\toprule
 & 20 & 40 & 60 & 80 \\
\midrule
Raw actions & 86.7 & 67.6 & 51.5 & 41.8 \\
VAE skills & 91.6 & 82.7 & 71.4 & 62.2 \\
\bottomrule
\end{tabular}
\end{table}

Table~\ref{tab:antmaze:success_rate_vs_distance} reports success rates on AntMaze for each world model, and Table~\ref{tab:antmaze:mean_success_vs_distance} averages them over world models. With raw actions, all three world models succeed only for goals one cell away, and reach at most 1\% success at two cells or more. With VAE skills, all three world models succeed at every distance, with success decreasing from 71--90\% at one cell to 3--11\% at four cells.

\begin{table}[t]
\caption{Success rate (\%) vs.\ goal distance (maze cells) on AntMaze, planned with CEM. Mean $\pm$ std over 3 training seeds.}
\label{tab:antmaze:success_rate_vs_distance}
\centering
\footnotesize
\renewcommand{\arraystretch}{1.3}
\begin{tabular}{llcccc}
\toprule
 &  & 1 & 2 & 3 & 4 \\
\midrule
\multirow[t]{2}{*}{LeWM} & Raw actions & 62$\pm$2 & 1$\pm$0 & 0$\pm$0 & 0$\pm$0 \\
 & VAE skills & \textbf{89$\pm$4} & \textbf{47$\pm$5} & \textbf{25$\pm$1} & \textbf{11$\pm$1} \\
\midrule
\multirow[t]{2}{*}{PLDM} & Raw actions & 56$\pm$1 & 0$\pm$1 & 0$\pm$0 & 0$\pm$0 \\
 & VAE skills & \textbf{90$\pm$5} & \textbf{43$\pm$6} & \textbf{24$\pm$3} & \textbf{3$\pm$2} \\
\midrule
\multirow[t]{2}{*}{SMWM} & Raw actions & 28$\pm$2 & 0$\pm$0 & 0$\pm$0 & 0$\pm$0 \\
 & VAE skills & \textbf{71$\pm$1} & \textbf{24$\pm$6} & \textbf{14$\pm$3} & \textbf{3$\pm$2} \\
\bottomrule
\end{tabular}
\end{table}

\begin{table}[t]
\caption{Mean success rate (\%) vs.\ goal distance (maze cells) on AntMaze, averaged over the three world models of Table~\ref{tab:antmaze:success_rate_vs_distance}.}
\label{tab:antmaze:mean_success_vs_distance}
\centering
\footnotesize
\renewcommand{\arraystretch}{1.3}
\begin{tabular}{lcccc}
\toprule
 & 1 & 2 & 3 & 4 \\
\midrule
Raw actions & 48.7 & 0.4 & 0.0 & 0.0 \\
VAE skills & 83.2 & 37.9 & 21.0 & 5.9 \\
\bottomrule
\end{tabular}
\end{table}

\subsection{Planner ablation}

\begin{figure*}[ht]
\centering
\includegraphics[width=\linewidth]{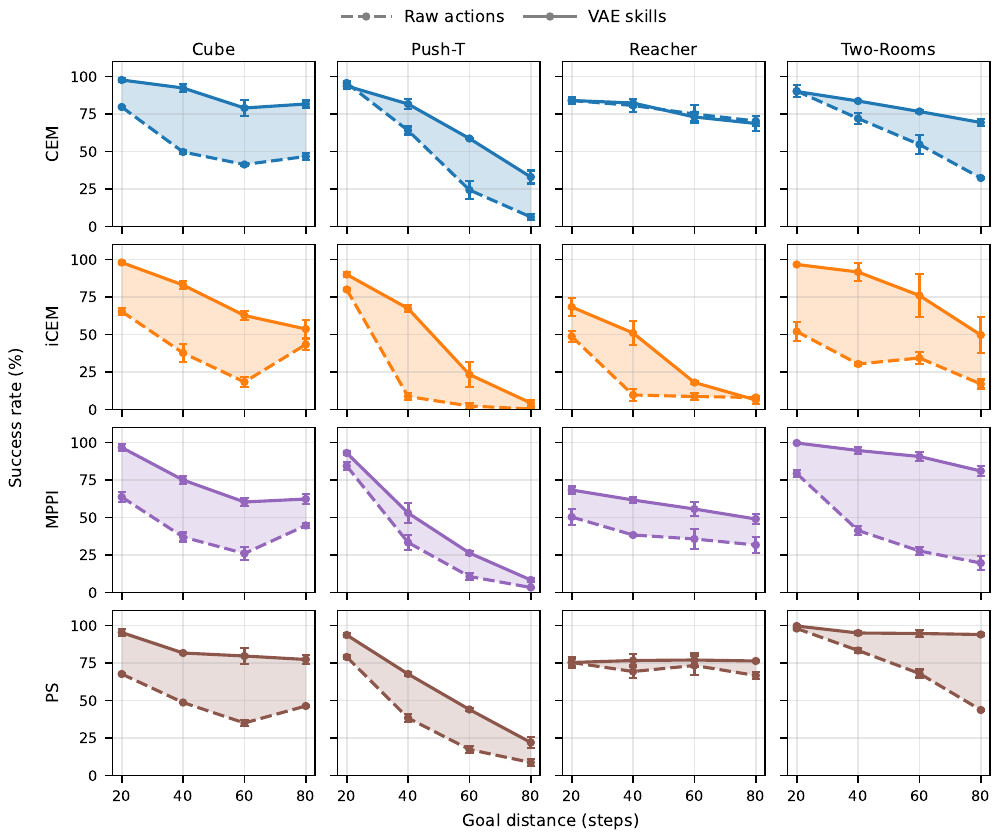}
\caption{Success rate vs.\ goal distance for raw actions and VAE skills with LeWM, across tasks and sampling-based planners.}
\label{fig:lewm_planner_ablation/success_rate_vs_distance}
\end{figure*}

Table~\ref{tab:lewm:planner_ablation:success_rate_vs_distance} and Figure~\ref{fig:lewm_planner_ablation/success_rate_vs_distance} report success rates with LeWM for the four sampling-based planners described in the main paper: CEM, iCEM, MPPI, and Predictive Sampling (PS). The world models and skill models are the same for all planners; only the planner changes. The CEM rows are the LeWM rows of Table~\ref{tab:methods:all:success_rate_vs_distance}. Averaged over goal distances, VAE skills have a higher success rate than raw actions in 15 of the 16 task and planner pairs, the exception being Reacher with CEM. At 80 steps, VAE skills are ahead in 14 of 16 pairs; the exceptions are Reacher with CEM (69\% vs.\ 70\%) and Reacher with iCEM (6\% vs.\ 8\%). With raw actions, iCEM and MPPI perform worse than CEM on most tasks. With VAE skills, the spread across planners on Two-Rooms narrows from 40 to 17 points (averaged over goal distances). Predictive Sampling, which only selects the best of the initial samples, is the best or close to the best planner with VAE skills at 80 steps on Cube, Reacher, and Two-Rooms (77\%, 76\%, and 94\%). Since PS relies on the initial distribution alone, this is consistent with Proposition~1. All planners use the same number of model evaluations per replanning step ($K \cdot I = 9000$).

\begin{table*}[ht]
\caption{Success rate (\%) vs.\ goal distance (environment steps) for raw actions and VAE skills with LeWM, across planners. Mean $\pm$ std over 3 training seeds.}
\label{tab:lewm:planner_ablation:success_rate_vs_distance}
\centering
\footnotesize
\renewcommand{\arraystretch}{1.3}
\begin{tabular}{llcccc}
\toprule
 &  & cube (20\,/\,40\,/\,60\,/\,80) & pusht (20\,/\,40\,/\,60\,/\,80) & reacher (20\,/\,40\,/\,60\,/\,80) & tworoom (20\,/\,40\,/\,60\,/\,80) \\
\midrule
\multirow[t]{2}{*}{CEM} & raw & 80$\pm$1\,/\,50$\pm$2\,/\,41$\pm$1\,/\,47$\pm$2 & \textbf{96$\pm$2}\,/\,64$\pm$3\,/\,24$\pm$6\,/\,6$\pm$2 & \textbf{84$\pm$2}\,/\,81$\pm$4\,/\,\textbf{75$\pm$6}\,/\,\textbf{70$\pm$3} & \textbf{90$\pm$4}\,/\,72$\pm$4\,/\,55$\pm$6\,/\,32$\pm$1 \\
 & vae & \textbf{98$\pm$1}\,/\,\textbf{92$\pm$3}\,/\,\textbf{79$\pm$6}\,/\,\textbf{82$\pm$2} & 94$\pm$2\,/\,\textbf{82$\pm$4}\,/\,\textbf{59$\pm$1}\,/\,\textbf{33$\pm$4} & \textbf{84$\pm$2}\,/\,\textbf{82$\pm$3}\,/\,73$\pm$3\,/\,69$\pm$5 & 90$\pm$0\,/\,\textbf{84$\pm$1}\,/\,\textbf{77$\pm$1}\,/\,\textbf{69$\pm$3} \\
\cline{1-6}
\multirow[t]{2}{*}{iCEM} & raw & 65$\pm$3\,/\,38$\pm$6\,/\,18$\pm$3\,/\,43$\pm$4 & 80$\pm$1\,/\,9$\pm$3\,/\,2$\pm$2\,/\,0$\pm$1 & 49$\pm$4\,/\,10$\pm$4\,/\,9$\pm$3\,/\,\textbf{8$\pm$1} & 52$\pm$7\,/\,30$\pm$2\,/\,34$\pm$4\,/\,17$\pm$3 \\
 & vae & \textbf{98$\pm$1}\,/\,\textbf{83$\pm$3}\,/\,\textbf{63$\pm$3}\,/\,\textbf{54$\pm$6} & \textbf{90$\pm$2}\,/\,\textbf{67$\pm$2}\,/\,\textbf{23$\pm$8}\,/\,\textbf{4$\pm$2} & \textbf{68$\pm$6}\,/\,\textbf{51$\pm$8}\,/\,\textbf{18$\pm$1}\,/\,6$\pm$3 & \textbf{97$\pm$1}\,/\,\textbf{92$\pm$6}\,/\,\textbf{76$\pm$14}\,/\,\textbf{50$\pm$12} \\
\cline{1-6}
\multirow[t]{2}{*}{MPPI} & raw & 64$\pm$3\,/\,37$\pm$3\,/\,26$\pm$4\,/\,45$\pm$2 & 84$\pm$3\,/\,33$\pm$5\,/\,11$\pm$3\,/\,3$\pm$1 & 50$\pm$5\,/\,38$\pm$1\,/\,36$\pm$7\,/\,32$\pm$6 & 79$\pm$3\,/\,41$\pm$3\,/\,28$\pm$3\,/\,20$\pm$5 \\
 & vae & \textbf{97$\pm$2}\,/\,\textbf{75$\pm$3}\,/\,\textbf{60$\pm$3}\,/\,\textbf{62$\pm$3} & \textbf{93$\pm$1}\,/\,\textbf{53$\pm$7}\,/\,\textbf{26$\pm$2}\,/\,\textbf{8$\pm$2} & \textbf{68$\pm$2}\,/\,\textbf{62$\pm$2}\,/\,\textbf{56$\pm$5}\,/\,\textbf{49$\pm$3} & \textbf{100$\pm$1}\,/\,\textbf{95$\pm$3}\,/\,\textbf{91$\pm$3}\,/\,\textbf{81$\pm$3} \\
\cline{1-6}
\multirow[t]{2}{*}{PS} & raw & 68$\pm$1\,/\,49$\pm$1\,/\,35$\pm$2\,/\,46$\pm$1 & 79$\pm$1\,/\,38$\pm$3\,/\,17$\pm$2\,/\,9$\pm$2 & \textbf{75$\pm$2}\,/\,69$\pm$5\,/\,73$\pm$6\,/\,67$\pm$3 & 98$\pm$1\,/\,83$\pm$2\,/\,68$\pm$3\,/\,44$\pm$1 \\
 & vae & \textbf{95$\pm$3}\,/\,\textbf{82$\pm$1}\,/\,\textbf{80$\pm$5}\,/\,\textbf{77$\pm$3} & \textbf{94$\pm$1}\,/\,\textbf{68$\pm$2}\,/\,\textbf{44$\pm$1}\,/\,\textbf{22$\pm$4} & \textbf{75$\pm$4}\,/\,\textbf{77$\pm$4}\,/\,\textbf{77$\pm$4}\,/\,\textbf{76$\pm$1} & \textbf{100$\pm$1}\,/\,\textbf{95$\pm$1}\,/\,\textbf{95$\pm$2}\,/\,\textbf{94$\pm$1} \\
\cline{1-6}
\bottomrule
\end{tabular}
\end{table*}

\subsection{Observation conditioning ablation}

In the main method, both the skill encoder $q_\theta(l \mid a_{t:t+k-1}, z_t)$ and the skill decoder $\hat a_\theta(l, z_t)$ are conditioned on the latent observation $z_t$. Table~\ref{tab:lewm:obs_conditioning_ablation:success_rate_vs_distance} compares this choice with variants where the encoder, the decoder, or both are not conditioned on $z_t$. In the row labels, \texttt{enc1} and \texttt{dec1} denote a conditioned encoder and decoder, and \texttt{enc0} and \texttt{dec0} denote unconditioned ones. This ablation uses LeWM on Push-T with a single training seed.
Conditioning both the encoder and the decoder gives the highest success rate at every goal distance, and the difference grows with distance (30\% vs.\ at most 8\% at 80 steps). Conditioning only one of the two performs similarly to conditioning neither.

\begin{table}[t]
\caption{Success rate (\%) vs.\ goal distance (environment steps) for VAE skills with LeWM on Push-T, with and without conditioning the skill encoder (\texttt{enc}) and decoder (\texttt{dec}) on the latent observation. Single training seed.}
\label{tab:lewm:obs_conditioning_ablation:success_rate_vs_distance}
\centering
\footnotesize
\renewcommand{\arraystretch}{1.3}
\begin{tabular}{llc}
\toprule
 &  & pusht (20\,/\,40\,/\,60\,/\,80) \\
\midrule
\multirow[t]{4}{*}{lewm} & vae dec1\_enc1 & \textbf{96}\,/\,\textbf{85}\,/\,\textbf{59}\,/\,\textbf{30} \\
 & vae dec1\_enc0 & 91\,/\,63\,/\,29\,/\,8 \\
 & vae dec0\_enc0 & 91\,/\,60\,/\,20\,/\,8 \\
 & vae dec0\_enc1 & 93\,/\,52\,/\,18\,/\,8 \\
\cline{1-3}
\bottomrule
\end{tabular}
\end{table}

\subsection{Positional embeddings ablation}

\begin{figure}[ht]
  \centering
  \includegraphics[width=\linewidth]{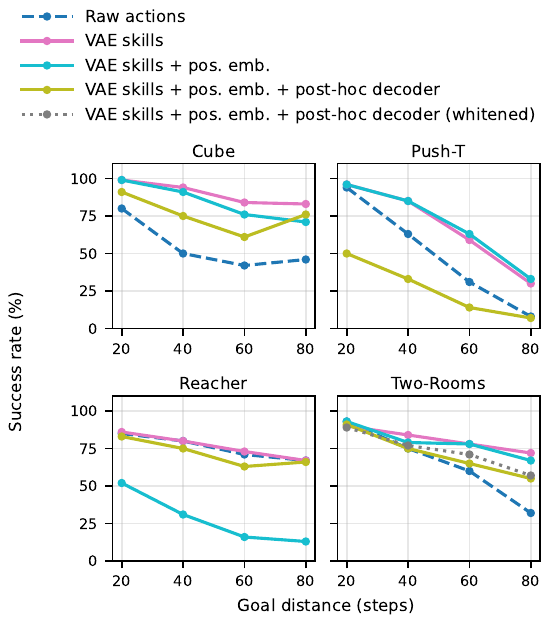}
  \caption{Success rate vs.\ goal distance for VAE skills with LeWM, with and without positional embeddings in the action encoder. Values as in Table~\ref{tab:lewm:pos_embedding:success_rate_vs_distance}.}
  \label{fig:lewm_pos_embedding/success_rate_vs_distance}
\end{figure}

In the main method, the skill encoder receives the $k$ actions of a chunk without positional embeddings. Since the encoder is a bidirectional Transformer, the skill is then invariant to the order of actions within the chunk. Table~\ref{tab:lewm:pos_embedding:success_rate_vs_distance} and Figure~\ref{fig:lewm_pos_embedding/success_rate_vs_distance} compare this choice with adding positional embeddings to the action encoder (\texttt{vae + pos}), and with adding positional embeddings and a post-hoc decoder (\texttt{vae + pos + posthocdec}). The post-hoc decoder uses the same architecture and losses, with a stop-gradient between the action encoder and the decoder, so the reconstruction loss does not reach the action encoder. This variant tests whether the skill latent spends capacity on action reconstruction at the expense of prediction. Raw actions are included as a reference. This ablation uses LeWM with a single training seed.

Positional embeddings do not improve over the main method on average. They give similar results on Push-T and Two-Rooms, lower results on Cube at 60 and 80 steps, and a large drop on Reacher (13\% vs.\ 67\% at 80 steps). The post-hoc decoder recovers most of this drop on Reacher (66\% at 80 steps), which suggests that, with positional embeddings, reconstruction takes up latent capacity needed for planning on this task. On Cube, Push-T, and Two-Rooms, the post-hoc decoder lowers success further, indicating that reconstruction gradients to the encoder are beneficial on these tasks.

\begin{table*}[ht]
\caption{Success rate (\%) vs.\ goal distance (environment steps) for VAE skills with LeWM, with and without positional embeddings in the action encoder, and with a post-hoc decoder. Single training seed.}
\label{tab:lewm:pos_embedding:success_rate_vs_distance}
\centering
\footnotesize
\renewcommand{\arraystretch}{1.3}
\begin{tabular}{llcccc}
\toprule
 &  & cube (20\,/\,40\,/\,60\,/\,80) & pusht (20\,/\,40\,/\,60\,/\,80) & reacher (20\,/\,40\,/\,60\,/\,80) & tworoom (20\,/\,40\,/\,60\,/\,80) \\
\midrule
\multirow[t]{5}{*}{lewm} & raw & 80\,/\,50\,/\,42\,/\,46 & 94\,/\,63\,/\,31\,/\,8 & 85\,/\,\textbf{80}\,/\,71\,/\,\textbf{67} & \textbf{93}\,/\,75\,/\,60\,/\,32 \\
 & vae & \textbf{99}\,/\,\textbf{94}\,/\,\textbf{84}\,/\,\textbf{83} & \textbf{96}\,/\,\textbf{85}\,/\,59\,/\,30 & \textbf{86}\,/\,\textbf{80}\,/\,\textbf{73}\,/\,\textbf{67} & 90\,/\,\textbf{84}\,/\,\textbf{78}\,/\,\textbf{72} \\
 & vae + pos & \textbf{99}\,/\,91\,/\,76\,/\,71 & \textbf{96}\,/\,\textbf{85}\,/\,\textbf{63}\,/\,\textbf{33} & 52\,/\,31\,/\,16\,/\,13 & \textbf{93}\,/\,79\,/\,\textbf{78}\,/\,67 \\
 & vae + pos + posthocdec & 91\,/\,75\,/\,61\,/\,76 & 50\,/\,33\,/\,14\,/\,7 & 83\,/\,75\,/\,63\,/\,66 & 91\,/\,75\,/\,65\,/\,55 \\
\cline{1-6}
\bottomrule
\end{tabular}
\end{table*}

\end{document}